\documentclass{article}

\usepackage{PRIMEarxiv}

\usepackage[utf8]{inputenc} 
\usepackage[T1]{fontenc}    
\usepackage{hyperref}       
\usepackage{url}            
\usepackage{graphicx}
\usepackage{amsmath}
\usepackage{booktabs}
\usepackage{hyperref}
\usepackage{algorithm,algorithmic}
\usepackage{subcaption}  
\usepackage[font=small,labelfont=bf]{caption}
\usepackage{amssymb} 
\newcommand{\cmark}{\checkmark}
\newcommand{\xmark}{\text{\sffamily X}}

\hypersetup{colorlinks=true, allcolors=blue}
\title{Bio-Inspired Robotic Houbara: From Development to Field Deployment for Behavioral Studies}

\author{Lyes Saad Saoud\textsuperscript{1} and Irfan Hussain\textsuperscript{1}\thanks{
\textsuperscript{1}Khalifa University Center for Autonomous Robotic Systems (KUCARS), Khalifa University, Abu Dhabi, United Arab Emirates.\\
\textbf{Citation:} Saad Saoud, L. and Hussain, I. *Bio-Inspired Robotic Houbara: From Development to Field Deployment for Behavioral Studies.* DOI:000000/11111.\\
}}

\begin{document}
\maketitle

\begin{abstract}
Biomimetic intelligence and robotics are transforming field ecology by enabling lifelike robotic surrogates that interact naturally with animals under real-world conditions. Studying avian behavior in the wild remains challenging due to the need for highly realistic morphology, durable outdoor operation, and intelligent perception that can adapt to uncontrolled environments. 
We present a next-generation bio-inspired robotic platform that replicates the morphology and visual appearance of the female Houbara bustard to support controlled ethological studies and conservation-oriented field research. The system introduces a fully \emph{digitally replicable fabrication workflow} that combines high-resolution structured-light 3D scanning, parametric CAD modelling, articulated 3D printing, and photorealistic UV-textured vinyl finishing to achieve anatomically accurate and durable robotic surrogates. A six-wheeled rocker--bogie chassis ensures stable mobility on sand and irregular terrain, while an embedded NVIDIA Jetson module enables real-time RGB and thermal perception, lightweight YOLO-based detection, and an autonomous visual servoing loop that aligns the robot’s head toward detected targets without human intervention. A lightweight thermal–visible fusion module enhances perception in low-light conditions. 
Field trials in desert aviaries demonstrated reliable real-time operation at 15–22\,FPS with sub-100\,ms latency and confirmed that the platform elicits natural recognition and interactive responses from live Houbara bustards under harsh outdoor conditions. This integrated framework advances biomimetic field robotics by uniting reproducible digital fabrication, embodied visual intelligence, and ecological validation, providing a transferable blueprint for animal–robot interaction research, conservation robotics, and public engagement.
\end{abstract}


\section{Introduction}

\begin{figure*}[t]
\centering
\includegraphics[width=0.7\linewidth]{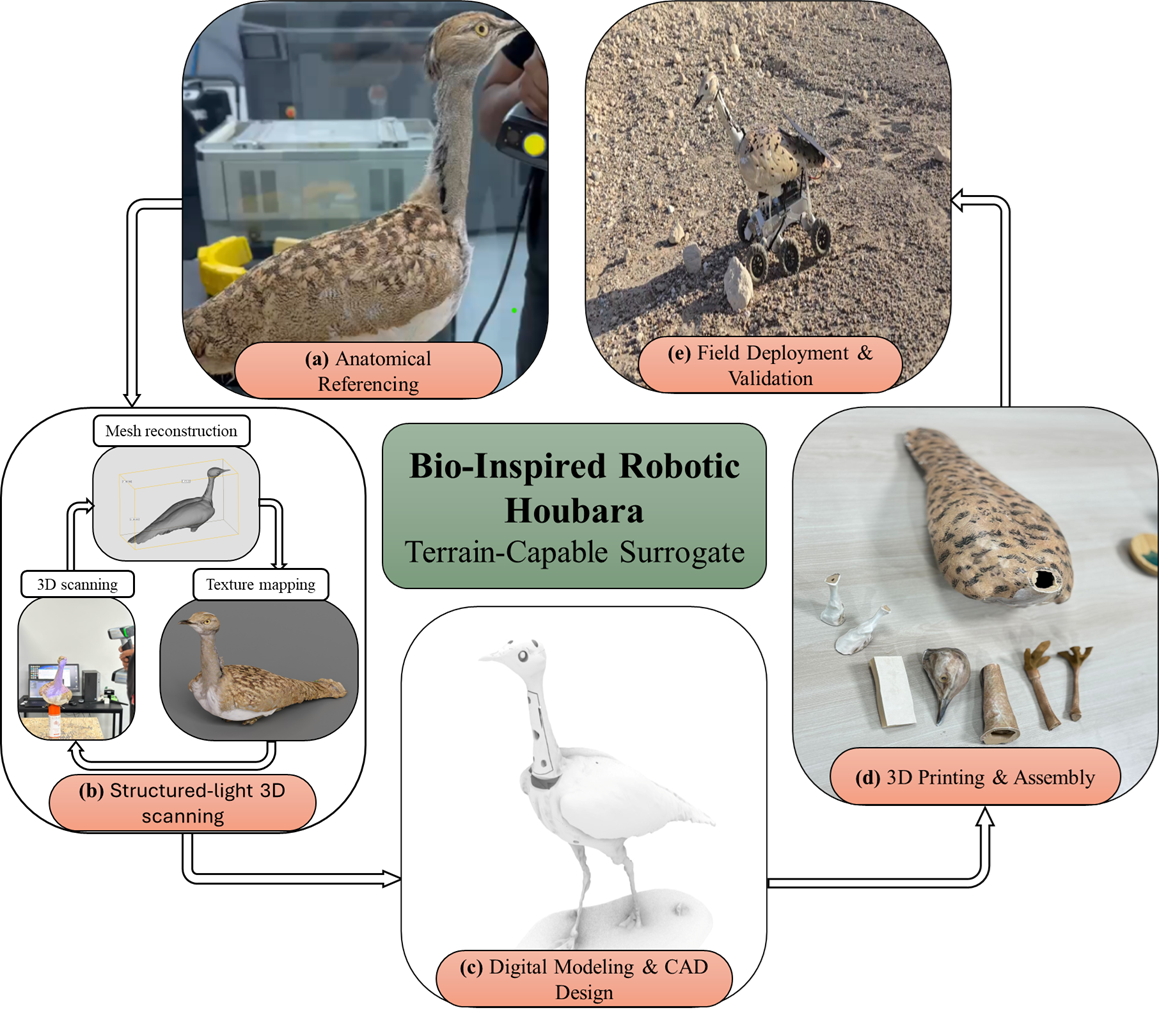}
\caption{Development pipeline of the bio-inspired Houbara robotic surrogate.\\
\textbf{(a)} Anatomical referencing of a preserved Houbara bustard specimen to capture accurate proportions and external morphology.\\
\textbf{(b)} Structured-light 3D scanning and texture mapping to generate a watertight, high-fidelity digital mesh.\\
\textbf{(c)} Computer-aided design (CAD) optimization of a modular articulated shell for field deployment and internal component integration.\\
\textbf{(d)} Three-dimensional printing and assembly followed by surface finishing with UV-textured vinyl for lifelike appearance and environmental robustness.\\
\textbf{(e)} Final field deployment and ecological validation of the terrain-capable robotic surrogate in natural desert habitats.}
\label{fig:main_pipeline}
\end{figure*}

Bio-inspired robots that replicate animal morphology and visual appearance are increasingly valuable in ecological monitoring, behavioral science, and conservation technology~\cite{bai2022amphibious}. They provide programmable, ethically controlled alternatives to live animals, enabling standardized and repeatable experiments without the welfare and logistical challenges of using real specimens~\cite{polverino2021controlling}. Animal–robot interaction (ARI) has been applied to diverse taxa, from influencing collective motion in insects~\cite{bonnet2017biohybrid} to eliciting social and courtship behaviors in vertebrates.

The integration of artificial intelligence (AI) has expanded ARI by enabling real-time perception and adaptive responses in unstructured field environments~\cite{SAADSAOUD2024102893}. However, bird-focused ARI remains less developed compared with aquatic and insect systems. Birds present unique challenges: complex feather layering and subtle visual cues are critical for conspecific recognition, while harsh outdoor habitats demand robust terrain mobility and resilience to heat, sand, and intense illumination~\cite{li2025morphing}. Achieving lifelike external appearance is particularly important because many bird species rely heavily on fine-scale color, texture, and reflectance for social signaling and recognition.

Earlier avian surrogates often relied on rigid constructs or taxidermy mounts that lacked durability and expressive articulation~\cite{biswal2024flapping}. Even when basic motion was introduced, these systems rarely achieved ecological acceptance or long-term outdoor performance. Advances in digital fabrication now enable the combination of anatomical precision with field-ready design: structured-light scanning preserves fine morphological detail; parametric CAD modeling supports modular shells with integrated actuation pathways; and UV-based texture transfer onto vinyl surfaces achieves reproducible, photorealistic coloration superior to manual painting~\cite{lee2019texture,shukla2023snake}.

\textbf{Beyond previous prototypes.} Our earlier HuBot platform~\cite{SAADSAOUD2025102939} demonstrated that a Houbara-inspired mobile robot could be deployed in desert habitats to observe and interact with live individuals without causing disturbance. While effective for initial behavioral studies, it depended on a preserved specimen, lacked a fully digital fabrication workflow, and provided no perception-driven head movement or thermal vision. The present work advances this foundation by (i) introducing a complete digital pipeline from structured-light scanning to UV-textured 3D-printed shells, (ii) integrating an on-board visual intelligence system with autonomous neck tracking and real-time perception–action coupling, and (iii) adding lightweight thermal–visible fusion and quantitative tracking performance analysis for day and low-light field operation.

\subsection*{Contributions}

The main technical contributions are:

\begin{itemize}
    \item \textbf{Digitally replicable biomimetic workflow:} A complete pipeline that transforms preserved bird specimens into lifelike robotic surrogates using structured light scanning, parametric CAD modeling, and articulated 3D printing.
    \item \textbf{High fidelity, scalable surface realism:} UV based texture transfer onto vinyl shells, providing durable, photorealistic finishes suitable for repeated outdoor use.
    \item \textbf{Terrain capable modular mobility:} A six wheel rocker bogie chassis and articulated head–neck mechanism that maintain stability on sand and irregular terrain while preserving external proportions.
    \item \textbf{Embodied visual intelligence:} An onboard NVIDIA Jetson platform that combines RGB and thermal sensing with TensorRT accelerated YOLO detection and a fully autonomous PID neck tracking loop operating in real time under field conditions.
    \item \textbf{Low light thermal and visible fusion:} A lightweight NightFusion method that provides interpretable heat guided enhancement and interactive temperature readout for nocturnal operation.
    \item \textbf{Ecological validation:} Controlled trials with captive Houbara bustards showing species specific recognition and approach behaviour in desert environments, indicating ecological acceptance and technical robustness.
    \item \textbf{Reproducibility and ethics:} Release of CAD models, meshes, firmware, and code together with an ethics statement covering approvals and animal care compliance to support transparent reuse and extension.
\end{itemize}

 \section{Related Work: Bio-Inspired Intelligence, Vision, and Autonomy in Field Robotics}

Bio-inspired robotic systems that replicate animal morphology, motion, and perceptual cues have become essential instruments for behavioral research and field robotics~\cite{bai2022amphibious,polverino2021controlling}. They enable controlled and repeatable experiments while reducing the ethical and logistical burdens of working with live animals. Rapid progress in actuation, sensing, and artificial intelligence has moved these platforms from laboratory prototypes to autonomous outdoor agents capable of adaptive interaction, real-time tracking, and long-duration deployment.

\begin{figure*}[t]
\centering
\includegraphics[width=0.9\linewidth]{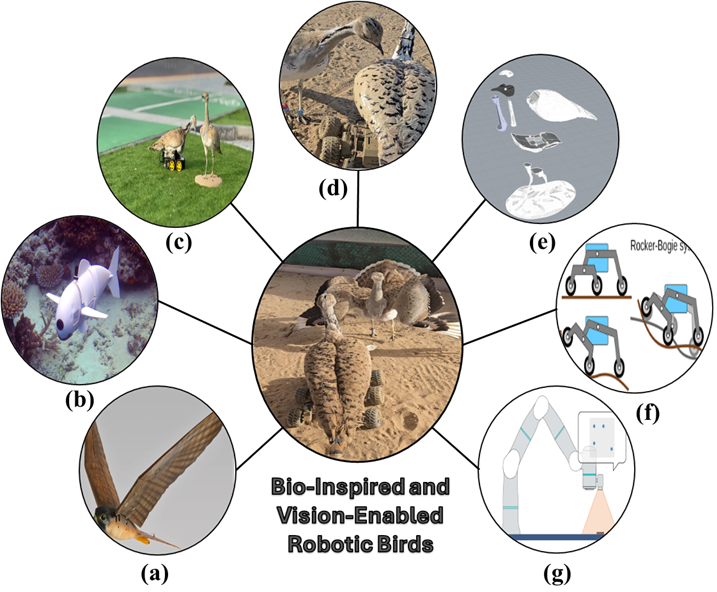}
\caption{Visual summary of seven domains shaping bio inspired and vision enabled robotic birds:
(a) recent advances in biomimetic bird robots and field AI;
(b) bio inspired robots in behavioral research;
(c) avian robotic systems for ecological and behavioral studies;
(d) vision enabled perception and animal–robot interaction;
(e) unified design principles for field ready robotic birds;
(f) terrain capable locomotion in bio inspired field robots; and
(g) control strategies for bio inspired or animal mimicking robots.
The central image shows a Houbara inspired robotic surrogate interacting with live conspecifics during desert field trials.}
\label{fig:bioinspired_circle}
\end{figure*}

This section surveys seven interrelated domains that shape next-generation avian-inspired robots (Figure~\ref{fig:bioinspired_circle}):
\noindent (i)~\textit{Recent advances in biomimetic bird robots and field AI}, emphasising lifelike morphology, embedded perception, and ecological use~\cite{wang2021survey,romano2025ari};
(ii)~\textit{Bio inspired robots in behavioral research}, demonstrating how morphology driven design supports reproducible ethological studies~\cite{polverino2021controlling,landgraf2016robofish};
(iii)~\textit{Avian robotic systems for ecological and behavioral studies}, addressing durability and anatomical fidelity~\cite{SAADSAOUD2025102939,li2025morphing,hoffmann2023bird};
(iv)~\textit{Vision enabled perception and animal–robot interaction}, detailing RGB thermal sensing and onboard control for interactive autonomy~\cite{kwon2025object,yi2023haffseg,song2023survey,fakrane2024human};
(v)~\textit{Toward unified design principles for field ready robotic birds}, integrating biomechanics, mechatronics, and digital fabrication~\cite{lee2019texture,shukla2023snake};
(vi)~\textit{Terrain capable locomotion in bio inspired field robots}, comparing Rocker–Bogie stability with legged and hybrid adaptability~\cite{senjaliya2022optimization,he2025probabilistic};
(vii)~\textit{Control strategies for bio inspired or animal mimicking robots}, including reinforcement learning, adaptive visual servoing, and simulation to reality transfer for perception driven autonomy~\cite{han2023spectral,jung2025ai}.

\subsection{Recent Advances in Biomimetic Bird Robots and Field AI}

Advances in biomimetic robotics and artificial intelligence are transforming wildlife research. Lifelike avian robots now enable controlled observation and interaction in natural habitats without disrupting typical behavior~\cite{wang2021survey}. Design efforts increasingly target accurate feather patterns, colouration, and postures to elicit species specific responses~\cite{romano2025ari,siddall2023ethorobotic}. Parallel results in animal–robot interaction show that morphologically faithful agents can shape and record behavior in ecologically valid ways~\cite{romano2025ari}. Beyond morphology, researchers combine robust outdoor mobility with energy efficient control for long duration operation, and hybrid configurations couple conventional vehicles with bio inspired surrogates to handle complex terrain~\cite{cloitre2012batoid}. Embedded artificial intelligence is central to these capabilities: convolutional and transformer architectures and lightweight YOLO variants now operate on edge devices for real time detection and tracking. Reinforcement learning and adaptive control improve robustness in unpredictable conditions~\cite{jung2025ai,ooko2025iot}. Thermal and multispectral sensing extend operation in low light and clutter. Open challenges include power management, data protection, and ethics, motivating lightweight and privacy aware models, standard multimodal datasets, and behavior aware control policies~\cite{fadzly2025future,ooko2025iot}.

\subsection{Bio Inspired Robots in Behavioral Research}

Bio inspired and biomimetic robots broaden the experimental toolkit for studying collective motion, predator prey interactions, and social signalling~\cite{polverino2021controlling}. Fish inspired platforms such as RoboFish demonstrate that realistic ocular features and naturalistic kinematics increase social acceptance in guppies and enable reproducible studies of group dynamics~\cite{landgraf2016robofish}. Robots integrated into zebrafish shoals, when driven by visual perception models, reproduce compatible collective patterns and enable interactive predator assays~\cite{cazenille2018mimetic,elkhoury2018interactive}. In insects, animal robot interaction has used morphology and motion cues to study local enhancement in the Mediterranean fruit fly, informing mechanisms of attraction relevant to pest management~\cite{romano2025ari}. Moving to avian field studies, \textit{HuBot} is a lifelike bird surrogate for non invasive observation and semi autonomous tracking in desert habitats~\cite{SAADSAOUD2025102939}. Complementary directions include soft robotics for safe contact and environmental adaptation, tensegrity inspired compliant structures, and robotic collectives that draw on biological swarming for robustness in challenging settings~\cite{sreedhar2025medical,wang2025intelligentsoft,hsieh2024tensegrity,zhao2025underwaterswarm}. Wall climbing adhesion extends mobility to vertical and variable substrates for ecological monitoring~\cite{pei2024wallclimbing}.

\subsection{Avian Robotic Systems for Ecological and Behavioral Studies}

Avian robotic systems have progressed far beyond static taxidermy. Contemporary platforms emphasise modularity and field readiness while preserving lifelike anatomical fidelity. Mechanical advances introduce morphing wings and compliant feather mechanisms that replicate natural flight dynamics and improve stability; bio informed flight control enhances manoeuvrability and energy economy during long outdoor missions~\cite{cheng2025rgblimp,wu2023remotesensing}. High fidelity morphology is central to ecological acceptance. Detailed digital modelling, bio inspired materials, and precise articulation enable reproduction of species specific motion patterns that support natural social signalling and courtship responses~\cite{patricelli2009robotics,patricelli2019robotics}. \textit{HuBot} illustrates how photorealistic external design and embedded perception can be combined for non invasive observation of elusive species in desert conditions~\cite{SAADSAOUD2025102939}. Durability remains essential: wind, sand, moisture, and vegetation require resilient structures and efficient actuation. Modular architectures and replaceable surface elements support maintenance and adaptation to experimental needs~\cite{wu2023remotesensing,cheng2025rgblimp}. Species specific surrogates, tuned in external features and behavior, have elicited grouping, mating, and avoidance responses, enabling reproducible and ethical field studies~\cite{frohnwieser2016robots,polverino2021controlling}.

\subsection{Vision Enabled Perception and Animal–Robot Interaction}

Progress in computer vision and embedded artificial intelligence now enables real time detection, robust tracking, and adaptive control in dynamic outdoor scenes. The fusion of visible and thermal imaging improves perception in low light, fog, and high contrast illumination. RGB thermal methods increase detection and tracking reliability and enhance environmental awareness across diverse habitats~\cite{kwon2025object,yi2023haffseg,song2023survey,fakrane2024human}. Hybrid adaptive fusion has improved segmentation accuracy and efficiency for mobile platforms~\cite{yi2023haffseg}. Visual servoing closes the loop between perception and motion, allowing precise alignment to detected targets and adaptive responses to unpredictable movement~\cite{reese2024optimizing,ivacko2024machine}. Edge platforms such as NVIDIA Jetson provide sufficient compute for onboard detection, tracking, and navigation while retaining compact and energy efficient form factors~\cite{choe2023run,choe2021benchmark,mohammadi2023caveline,zhang2024autonomous}. Autonomous bird robots that combine RGB thermal perception, visual servoing, and efficient embedded intelligence can track wildlife, respond to behavioral cues, and collect data with minimal human intervention~\cite{kwon2025object,song2023survey}. Key challenges include autonomous calibration and robust sensor fusion across modalities~\cite{yang2024autocalibration,brenner2023fusion} and sustained real time performance on energy constrained platforms.

\subsection{Toward Unified Design Principles for Field Ready Robotic Birds}

Ecologically valid robotic birds require an integrated design approach that unites biological fidelity with engineering robustness. Accurate morphology remains fundamental: body proportions, skeletal structure, and feather arrangement must support natural responses and stable flight. Biohybrid morphing wings such as those of PigeonBot highlight the importance of coordinated feathers~\cite{chang2020soft}. Large scale platforms such as HIT Hawk and HIT Phoenix couple aerodynamic modelling with advanced structures to enhance stability in outdoor conditions~\cite{pan2021hit}. Structured light scanning and UV based texture mapping allow precise replication of surface features and coloration for species recognition~\cite{hoffmann2023bird,fisher2012bioinspired}. Energy efficient actuation is equally important. Smart composites with embedded sensing and actuation mimic multifunctional avian tissues~\cite{hoffmann2023bird}; muscle like direct drive motors offer high force density for dynamic wing motion~\cite{ruddy2011muscle}; modular multibody wing systems provide efficient lift and thrust~\cite{sharma2022simplified}. Robust autonomy relies on vision based navigation and adaptive control for stable flight under environmental variation, supported by onboard planning and disturbance rejection~\cite{pan2021hit,yuan2014indoor}. Sustainable design, including self stiffening or deployable wings and environmentally friendly materials, increases resilience and reduces ecological impact in long term monitoring~\cite{rojas2024bioinspired,mazzolai2021towards,mazzolai2022advancing}.

\subsection{Terrain Capable Locomotion in Bio Inspired Field Robots}

Traversing unstructured terrain remains a major challenge for field robots. Among wheeled platforms, the Rocker–Bogie mechanism maintains passive stability and continuous ground contact on moderately rough terrain, yet performance degrades when obstacles are large or the substrate yields~\cite{menon2007adaptation}. Legged robots mitigate these limits by adjusting gait and posture to maintain traction on uneven or deformable surfaces~\cite{zhao2021terrain}. Terrain classification, force based adaptation, probabilistic feedback control, and fault tolerant gaits improve stability across sand, gravel, and vegetation~\cite{he2025probabilistic,asif2012improving}. Neural controllers and artificial hormone systems further enhance adaptability and energy economy~\cite{homchanthanakul2019neural}. Hybrid systems merge rolling efficiency with stepping adaptability. Designs such as OmniQuad provide omnidirectional motion with mecanum wheels and leg like elements~\cite{iotti2025omniquad}, while other wheeled and legged robots switch between modes across deserts, grasslands, and wetlands~\cite{wang2025wheeled}. Mechanical solutions such as four bar linkages increase stiffness without sacrificing mobility~\cite{banabic2024mobile}. Bio inspired strategies, including cockroach and centipede inspired bodies, central pattern generator control, morphological intelligence with active tails, and even multimodal morphologies that support walking, swimming, and flying, broaden applicability for ecological monitoring and search and rescue~\cite{baisch2014high,koh2010centipede,chen2014adaptive,siddall2021compliance,sun2025bipedal,polzin2025robotic}.

\subsection{Control Strategies for Bio Inspired or Animal Mimicking Robots}

Control strategies now emphasise reinforcement learning, adaptive visual servoing, and simulation to reality transfer to achieve perception driven autonomy. Deep reinforcement learning acquires locomotion and interaction skills in simulation and in real environments; hierarchical formulations improve energy use and stability~\cite{hameed2022tadpole,xiang2021task,azimi2025hierarchical}. Transfer pipelines that employ domain adaptation and cycle consistent translation improve reliability outside the simulator~\cite{jiang2022closedloop,shi2023sim2real,rao2020rlcyclegan}. Adaptive visual servoing combines neural estimators with online parameter updates to handle uncertain calibration and dynamic scenes, maintaining robust tracking in real time~\cite{han2023spectral,xu2020underwater,wang2019softvs}. Domain adaptation and data augmentation bridge synthetic training with field deployment, while embedded platforms execute detection and control in real time for responsive behavior in harsh environments~\cite{julian2020latent,liu2023digitaltwin,jung2025ai}. Together these approaches are moving bio inspired robots from proofs of concept to robust autonomous instruments for long term ecological studies and public engagement.

\section{Bio-Inspired Design Workflow}

\begin{figure*}[t]
\centering
\includegraphics[width=0.99\linewidth]{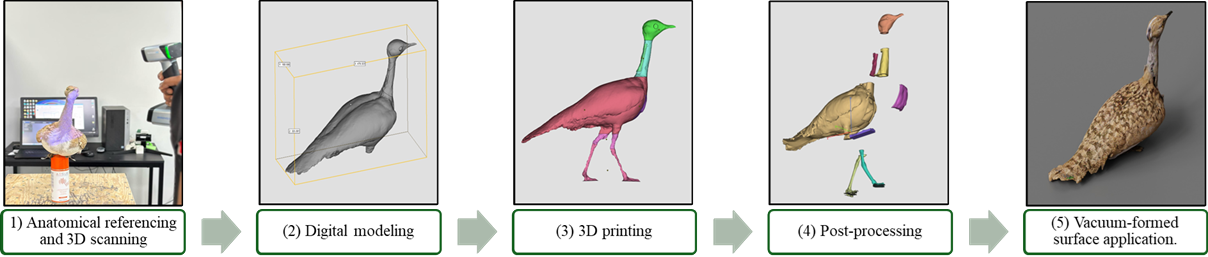}
\caption{Five-stage robotic design workflow.  
\textbf{(1)} Anatomical referencing and structured-light scanning of a preserved Houbara bustard under diffuse illumination.  
\textbf{(2)} Digital modeling and modular segmentation of the 3D mesh with integrated slots for actuators and electronics.  
\textbf{(3)} Additive manufacturing of articulated components with internal channels for wiring and motion.  
\textbf{(4)} Surface preparation with sanding, epoxy filling, and primer coating for durability and smoothness.  
\textbf{(5)} Application of weather-resistant photorealistic vinyl textures on a vacuum-formed shell to achieve lifelike appearance and outdoor robustness.}
\label{fig:design_pipeline}
\end{figure*}

The robotic Houbara bustard was developed through a structured five-stage pipeline that integrates anatomical digitization, computer-aided design, additive manufacturing, and photorealistic surface finishing (Figure~\ref{fig:design_pipeline}). The workflow was designed for reproducibility, modular upgrades, and field durability, ensuring that the platform can support both ecological experiments and public technology demonstrations.

\subsection*{Step~1: Anatomical referencing and 3D scanning}

A preserved female Houbara bustard specimen was digitized using structured-light scanning at sub-millimeter resolution under diffuse illumination. Complementary photogrammetry captured high-resolution color and fine plumage texture for later mapping. Combining geometric and photometric capture is a well-established approach in biomedical and zoological replication~\cite{moncayo2023implant}, ensuring both dimensional accuracy and faithful visual reproduction.

\subsection*{Step~2: Digital modeling and reverse engineering}

Raw scans were denoised, mesh-repaired, and topologically optimized in MeshLab, then reverse-engineered in Autodesk Fusion~360. The external shell was segmented into 14 interlocking anatomical modules, each with integrated cavities for servomotors, wiring, and control electronics (Figure~\ref{fig:digital_design}a). Parametric CAD modeling enabled precise joint articulation, efficient replacement of damaged sections, and future upgrades such as alternative actuation systems or additional sensing payloads.

\subsection*{Step~3: Additive manufacturing and assembly}

All structural parts were fabricated using fused filament fabrication (FFF) with 0.4~mm nozzles and PLA-based materials optimized for strength and heat stability. The CAD design incorporated overhang constraints, snap-fit joints, and embedded screw channels to minimize support material and enable tool-free assembly. This approach accelerated iteration cycles and reduced downtime during outdoor testing.

\subsection*{Step~4: Surface preparation and finish optimization}

Printed components were sanded, epoxy-filled at seams, and coated with high-adhesion primer to enhance mechanical durability and provide a smooth base for surface finishing. These practices, common in animatronics and museum-grade replicas, significantly improve resistance to UV degradation, high desert temperatures, and abrasive sand exposure.

\subsection*{Step~5: Texture mapping and photorealistic vinyl wrapping}

\begin{figure*}[t]
\centering
\begin{minipage}[t]{0.48\textwidth}
\centering
\includegraphics[width=\linewidth]{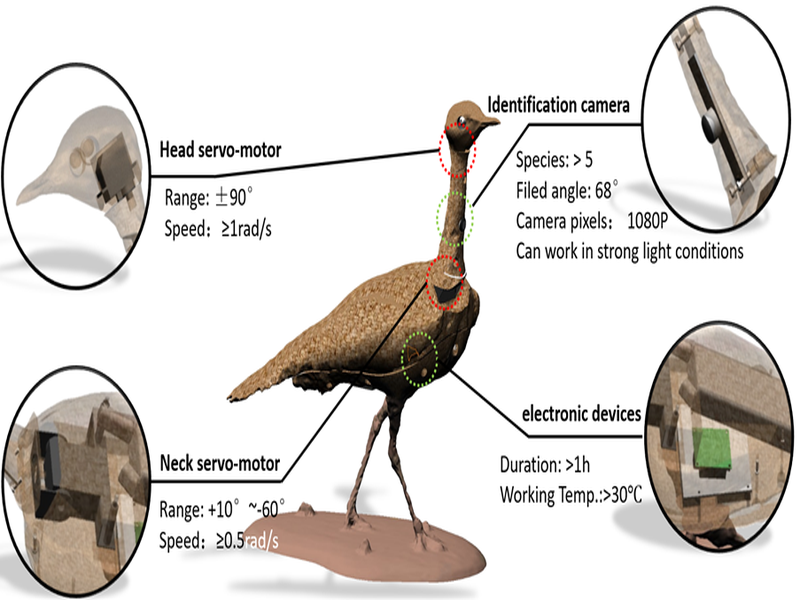}
\caption*{(a) Exploded CAD model}
\end{minipage}
\hfill
\begin{minipage}[t]{0.48\textwidth}
\centering
\includegraphics[width=\linewidth]{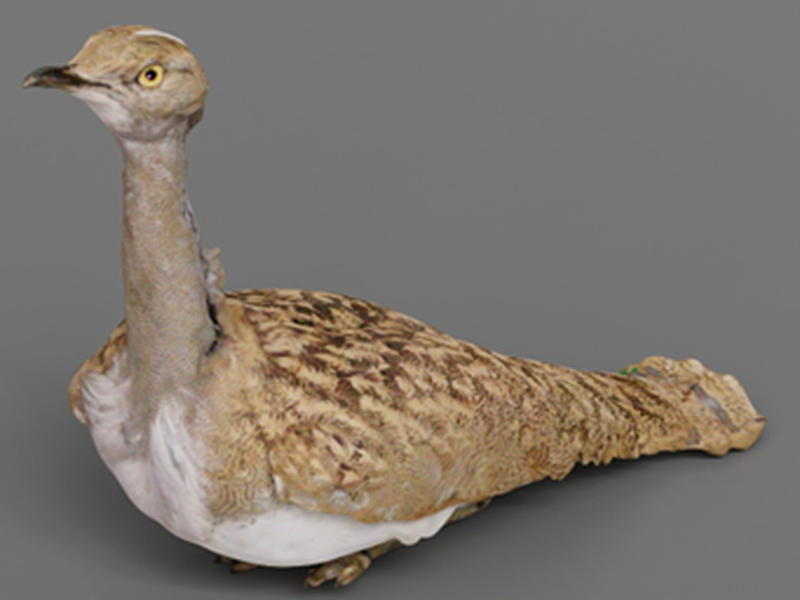}
\caption*{(b) UV texture map}
\end{minipage}
\caption{Digital design outputs. \textbf{(a)} Exploded CAD model with servo housings, electronics bays, and wiring channels. \textbf{(b)} Optimized UV texture map aligned to anatomical landmarks for accurate color reproduction in vinyl printing.}
\label{fig:digital_design}
\end{figure*}

The UV-unwrapped surface derived from the photogrammetric model was color-corrected and aligned to anatomical landmarks to avoid visible seams. High-resolution textures were solvent-printed on weather-resistant matte vinyl and heat-applied to the vacuum-formed shell. Compared with traditional hand painting, this method yielded superior color fidelity, reproducibility across multiple units, and reduced glare under strong sunlight (Figure~\ref{fig:digital_design}b).

\subsection*{Final integration and field validation}

\begin{figure}[t]
\centering
\includegraphics[width=0.48\textwidth]{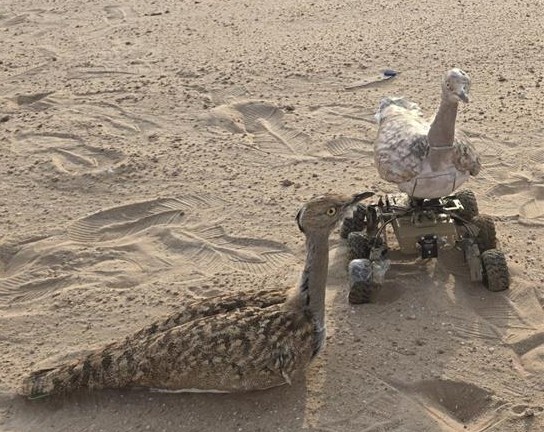}
\caption{Completed robotic shell with photogrammetric vinyl wrapping. The matte finish enhances realism, reduces glare in desert sunlight, and withstands UV, heat, and abrasive wear during field operation.}
\label{fig:sticker_comparison}
\end{figure}

The fully assembled matte-wrapped shell (Figure~\ref{fig:sticker_comparison}) was field-tested in desert conditions. Trials confirmed high resilience to UV exposure, abrasive sand, and elevated temperatures while maintaining compatibility with RGB and thermal cameras for day–night operation. Compared with earlier hand-painted prototypes, the vinyl-based workflow improved reproducibility, reduced fabrication time, and extended service life during long-term ecological deployments.

\section{Thermal–Visible Fusion for Low-Light Monitoring}

To ensure reliable perception under poor illumination and at night, we integrate a lightweight thermal–visible fusion method, termed NightFusion. An RGB camera is paired with a FLIR Lepton long-wave infrared (LWIR) module (8–14~$\mu$m) providing radiometric temperature readings. 

Thermal frames are first resized to the RGB resolution and robustly normalized to $[0,1]$ using lower and upper percentiles $(p_\ell,p_h)$. A coarse illumination proxy
\[
L = \big[\text{stretch}(T_{01};p_\ell,p_h)\big]^{\gamma}
\]
is smoothed by a small Gaussian kernel and refined using an edge-aware guided filter. Temporal stabilization is achieved with an exponential moving average
\[
\hat{L}_t = a\,\hat{L}_{t-1} + (1-a)\,\tilde{L}_t,\quad a\in[0,0.98],
\]
reducing flicker without lagging transient heat sources. 

Fusion proceeds by gain modulation of the visible frame
\[
G = \alpha + \beta\,\hat{L}_t,\qquad I_f = \operatorname{clip}(I\odot G,0,1),
\]
where $\alpha$ is the base gain and $\beta$ controls thermal guidance strength. After gain modulation, we enhance detail by an unsharp mask and optionally apply CLAHE on the Y channel before converting back to BGR. The approach is single-pass and runs in real time on CPU.

\textbf{Implementation details.}
We used $p_\ell=2$, $p_h=98$, $\gamma=0.7$, $a=0.9$, Gaussian kernel $k=7$, $\alpha=0.7$, $\beta=1.6$, unsharp strength $0.5$, and CLAHE clip $2.0$ with an $8\times8$ grid. Guided filtering uses OpenCV’s \texttt{ximgproc.guidedFilter} when available and a fast guided filter fallback otherwise. Radiometric temperature readouts from the Lepton SDK are mapped to a GUI for pixel-level inspection.

\section{Mobility and Physical Platform Evolution}

\begin{figure}[t]
\centering
\includegraphics[width=0.98\linewidth]{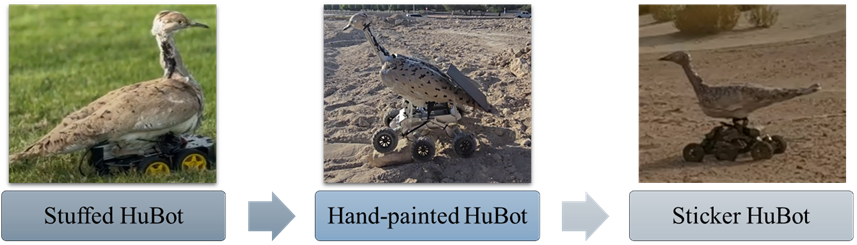}
\caption{Mechanical evolution of the HuBot robotic surrogate. \textbf{Left:} static taxidermy-based prototype with limited mobility and outdoor resilience. \textbf{Center:} six-wheeled rocker--bogie chassis with a 3D-printed, hand-painted shell offering improved robustness but limited visual consistency. \textbf{Right:} final field-ready platform with a vinyl-wrapped PLA shell using photogrammetrically derived UV textures, enhancing durability and day--night sensing for extended desert deployment.}
\label{fig:version_evaluation}
\end{figure}

The mechanical evolution of the HuBot platform was guided by the need to balance anatomical fidelity, ecological robustness, and reliable terrain traversal. As illustrated in Figure~\ref{fig:version_evaluation}, the design progressed through three major iterations, each addressing limitations identified during earlier deployments.

The initial prototype combined a static taxidermy mount with a simple four-wheel chassis. While visually realistic for behavioral trials, it lacked mobility and structural durability under outdoor conditions, restricting its use to controlled aviaries and short outdoor sessions.

To overcome these shortcomings, the second version adopted a six-wheeled rocker--bogie base inspired by planetary rover architectures~\cite{yao2024staf, garcia2024rolling}. This configuration improved stability and sand mobility, while replacing the fragile taxidermy body with a 3D-printed polylactic acid (PLA) shell. The new structure was modular and more resilient but relied on hand painting, leading to inconsistent finishes and limited reproducibility when multiple units were required.

The final iteration introduced a vinyl-wrapped PLA shell with UV-mapped textures derived from high-resolution photogrammetric scans of preserved Houbara specimens. This approach significantly improved surface fidelity, durability, and manufacturing repeatability, enabling consistent production of field-ready units. The matte vinyl reduced glare, enhancing the performance of the onboard RGB and thermal cameras for day--night operation. The rocker--bogie chassis was retained to maintain stable locomotion across dunes and irregular desert terrain, supporting long-duration deployments.

\section{Embedded Intelligence and Vision–Servo Architecture}

The embedded control system of HuBot was designed to enable real-time perception, adaptive actuation, and modular field operation under outdoor desert conditions. At its core is an NVIDIA Jetson Orin~NX (8~GB or 16~GB), capable of up to 100~INT8~TOPS, mounted on a reComputer~J4012 carrier running JetPack~6.1. This configuration supports TensorRT-accelerated inference for lightweight object detectors such as YOLOv11n, fine-tuned on a custom dataset of Houbara and surrogate classes collected under diverse lighting and environmental scenarios to ensure robust real-time detection during interaction trials.

The perception module combines RGB and thermal modalities: a daylight RGB camera and a PureThermal~3 (PT3) module with a FLIR Lepton core for low-light and night-vision imaging. The PT3 provides radiometric thermal data to the acquisition pipeline, though the current visual servo loop relies on the RGB feed.

\begin{figure*}[t]
\centering
\begin{minipage}[t]{0.58\textwidth}
\centering
\includegraphics[width=\linewidth]{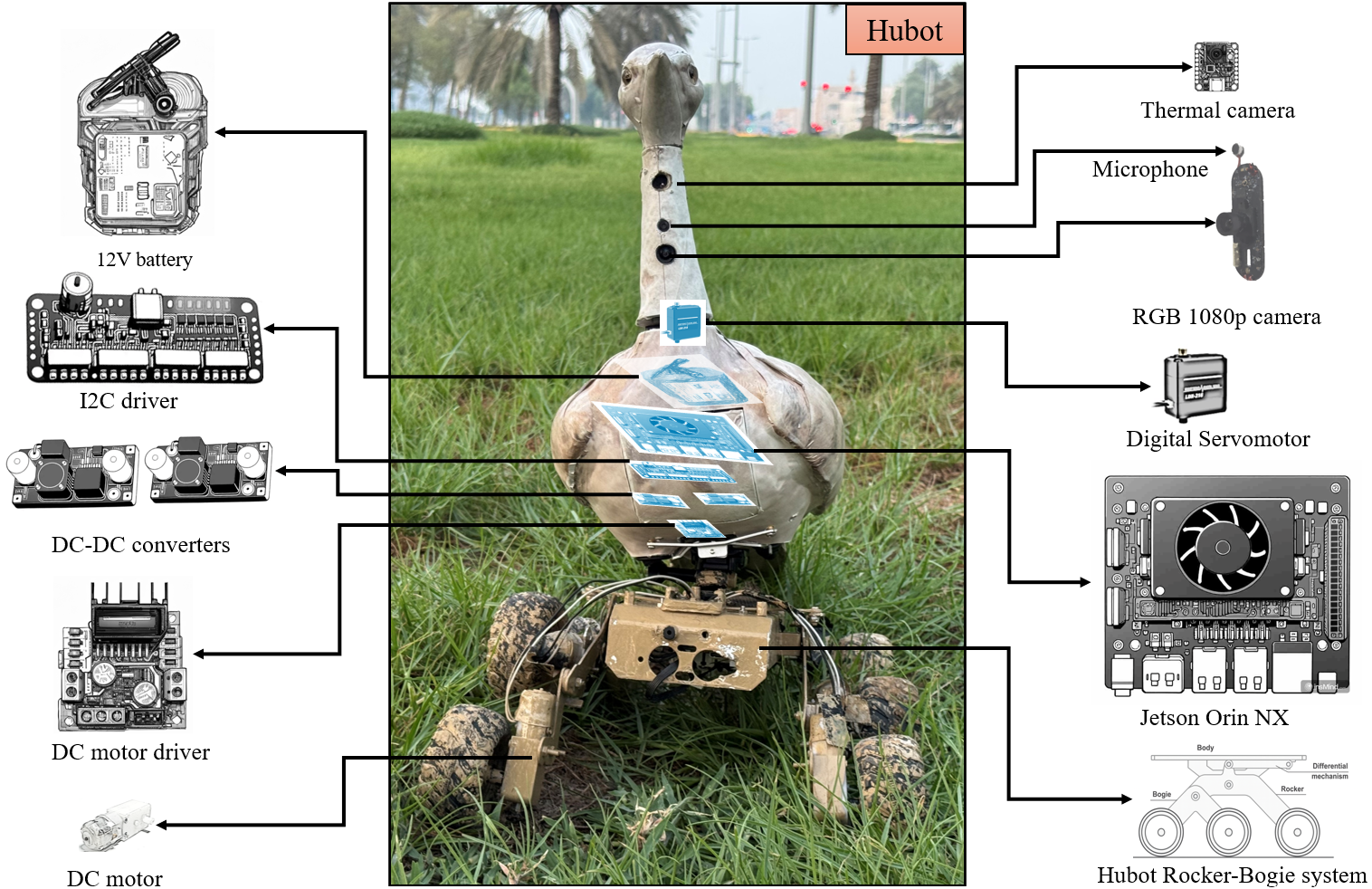}
\caption*{(a) System architecture}
\end{minipage}
\hfill
\begin{minipage}[t]{0.38\textwidth}
\centering
\includegraphics[width=\linewidth]{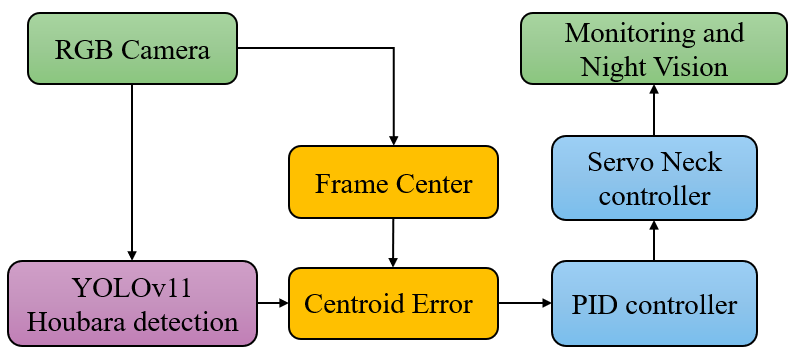}
\caption*{(b) Visual tracking loop}
\end{minipage}
\caption{Embedded intelligence and vision–servo control of the HuBot platform. \textbf{(a)} System architecture integrating onboard compute, dual visual sensing, PID-driven pan–tilt actuation, rocker--bogie mobility, power distribution, and GUI-based supervision. \textbf{(b)} Real-time RGB perception feeding YOLOv11n detection; centroid errors are processed by a PID loop to orient the neck via PCA9685-driven servos. The thermal camera supports night-vision logging but is not yet included in the active control loop.}
\label{fig:system_and_tracking_architecture}
\end{figure*}

\begin{figure}[t]
\centering
\includegraphics[width=0.88\linewidth]{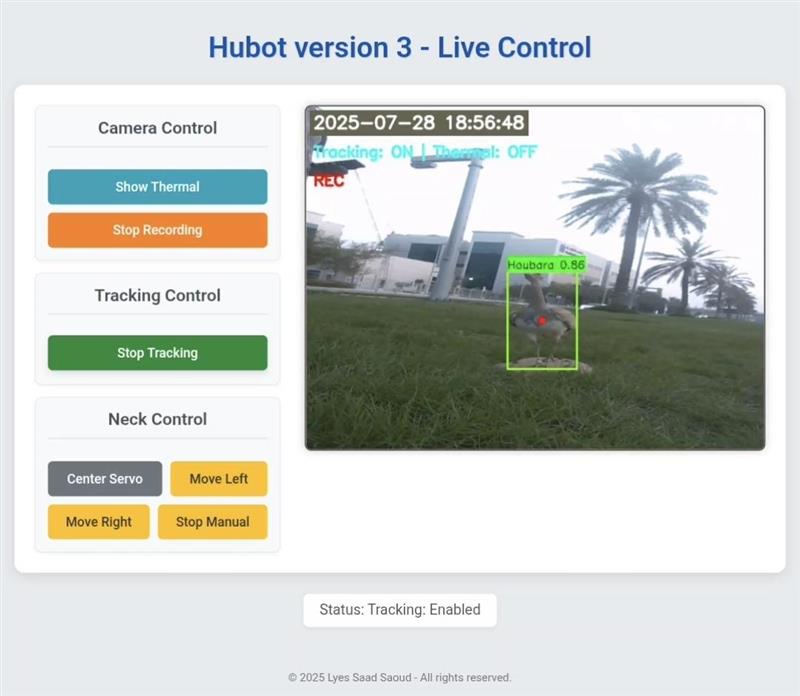}
\caption{Graphical user interface of HuBot during live operation. The interface provides camera controls (RGB/thermal toggle, recording), YOLO-based tracking activation, and pan–tilt neck control with on-frame overlays showing detection status, thermal inset, and servo angle readout.}
\label{fig:gui_interface}
\end{figure}

Visual tracking is user-activated via the platform’s graphical user interface and drives a pan–tilt neck mechanism actuated by servos through a PCA9685 module on the I\textsuperscript{2}C bus. Once a Houbara is detected by YOLOv11n, the horizontal error between the bounding-box centroid and the image center,
\[
e(t)=x_{\text{target}}(t)-x_{\text{center}},
\]
is mapped to a neck command using a proportional–integral–derivative (PID) controller (Algorithm~\ref{alg:pid_tracking}) to maintain continuous target alignment. The complete perception-to-actuation loop is shown in Figure~\ref{fig:system_and_tracking_architecture}(b), with a live operator interface in Figure~\ref{fig:gui_interface} and quantitative tracking accuracy in Figure~\ref{fig:tracking_accuracy}.

Mobility is provided by six brushed DC motors in a rocker--bogie configuration, driven by an L298N H-bridge using PWM from the Jetson’s GPIO. Power is supplied by an 11.1~V, 2200~mAh LiPo battery regulated by dual LM2596 buck converters: one for the compute unit and one for drive and control electronics.

System-wide telemetry, including synchronized video streams, detection metadata, tracking status, and battery voltage, is logged to an onboard USB SD card and displayed through a mobile or desktop graphical user interface (GUI). Table~\ref{tab:electronics_summary} summarizes the embedded hardware and their functions.

\begin{figure}[t]
\centering
\includegraphics[width=0.95\linewidth]{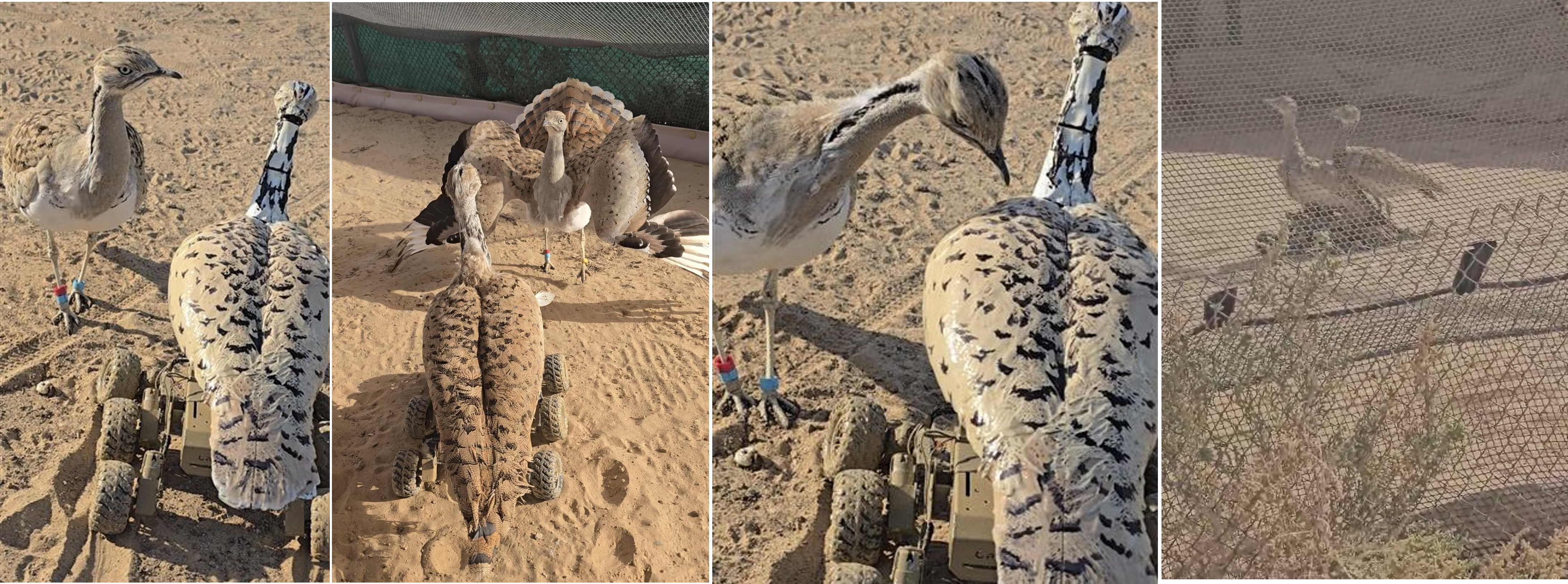}
\caption{Interaction with an adult male Houbara during field deployment. The articulated neck and photorealistic surface finish elicited species-specific courtship responses, confirming ecological acceptance under harsh outdoor conditions.}
\label{fig:real_interaction}
\end{figure}
\begin{algorithm}[t]
\caption{PID-based servo tracking activated through the GUI}
\label{alg:pid_tracking}
\begin{algorithmic}[1]
\STATE \textbf{Input:} Target centroid $x_{\text{target}}$, image center $x_{\text{center}}$
\STATE \textbf{Parameters:} PID gains $K_p$, $K_i$, $K_d$, time step $\Delta t$
\STATE Initialize: $e_{\text{prev}} \gets 0$, $e_{\text{sum}} \gets 0$, current servo angle $\theta$
\WHILE{camera active and tracking enabled}
    \STATE $e \gets x_{\text{target}} - x_{\text{center}}$
    \STATE $e_{\text{sum}} \gets e_{\text{sum}} + e \cdot \Delta t$
    \STATE $e_{\text{diff}} \gets (e - e_{\text{prev}}) / \Delta t$
    \STATE $\theta_{\text{new}} \gets \theta + K_p e + K_i e_{\text{sum}} + K_d e_{\text{diff}}$
    \STATE $\theta_{\text{new}} \gets \mathrm{clamp}(\theta_{\text{new}}, \theta_{\min}, \theta_{\max})$
    \STATE Move servo to $\theta_{\text{new}}$
    \STATE $e_{\text{prev}} \gets e$
\ENDWHILE
\end{algorithmic}
\end{algorithm}

\begin{table}[t]
\centering
\caption{Embedded hardware and functional modules of the HuBot platform}
\label{tab:electronics_summary}
\begin{tabular}{p{3.2cm} p{5.0cm} p{6.5cm}}
\toprule
\textbf{Component} & \textbf{Model} & \textbf{Function} \\
\midrule
Embedded processor & NVIDIA Jetson Orin NX & Onboard perception, inference, and control \\
RGB camera & UVC RGB module & Daylight visual sensing \\
Thermal camera & PureThermal~3 (FLIR Lepton) & Low-light and night-vision sensing \\
Object detector & YOLOv11n (TensorRT) & Real-time Houbara detection \\
Neck actuation & PCA9685 + PWM servos & PID-driven pan–tilt tracking \\
Mobility system & Six brushed DC motors + L298N & Rocker--bogie terrain navigation \\
Power converters & Dual LM2596 buck modules & Regulated power supply for compute and drive \\
Battery pack & 11.1~V, 2200~mAh LiPo & Field-suitable power source \\
Storage module & USB SD card & Logging of video, telemetry, and metadata \\
User interface & Mobile/desktop GUI & Live monitoring and manual override \\
\bottomrule
\end{tabular}
\end{table}


\section{Deployment, Interaction, and Vision-Based Tracking Validation}

The final HuBot prototype was evaluated in ecologically relevant settings, including indoor aviaries and outdoor desert enclosures. The study pursued two objectives: (i) assessing whether the robot elicits ethologically meaningful responses from live Houbara bustards, and (ii) validating the vision pipeline for continuous observation, low-light operation, and closed-loop tracking.

\textbf{Ecological validation with live Houbara.}
During the peak courtship season, the robotic surrogate was evaluated with 40 captive-bred Houbara bustards (\emph{Chlamydotis macqueenii}) housed individually in outdoor enclosures. Each bird was exposed in a single controlled trial comprising a baseline phase, a static taxidermic stimulus, and the robotic surrogate performing predefined movement patterns (Figure~\ref{fig:real_interaction}). Behavioral metrics included approach latency and distance, orientation and head alignment, vigilance, and courtship displays. 

Across these 40 trials, most birds approached the robot and oriented toward it, with several males performing typical courtship postures. Data were analyzed with within-subject nonparametric tests for continuous measures and paired proportion tests for categorical outcomes. These analyses supported that the platform was generally accepted as a conspecific-like agent under natural desert conditions. Throughout repeated sessions the robot maintained structural and visual fidelity despite temperatures exceeding 40\,$^\circ$C, direct sunlight, abrasive sand, and occasional physical contact.

\begin{figure*}[t]
\centering
\begin{subfigure}[b]{0.24\textwidth}
    \includegraphics[width=\linewidth]{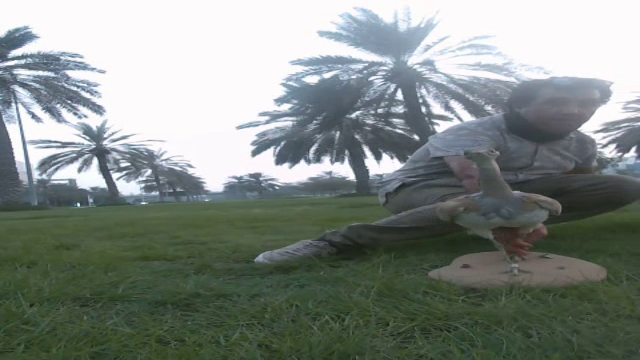}
    \caption*{\footnotesize RGB~1}
\end{subfigure}\hfill
\begin{subfigure}[b]{0.24\textwidth}
    \includegraphics[width=\linewidth]{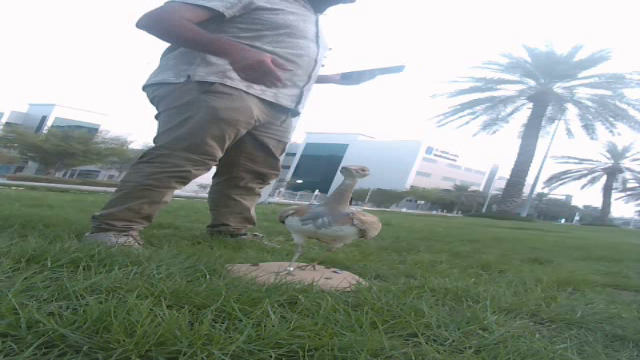}
    \caption*{\footnotesize RGB~2}
\end{subfigure}\hfill
\begin{subfigure}[b]{0.24\textwidth}
    \includegraphics[width=\linewidth]{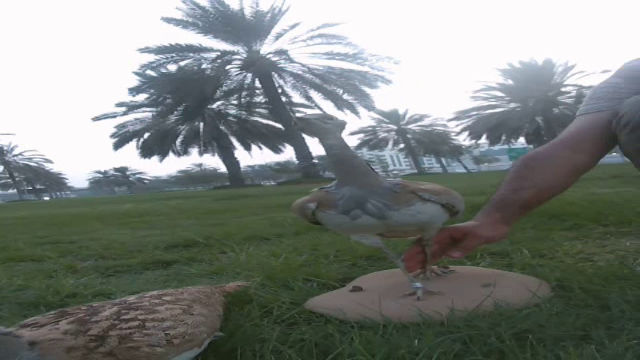}
    \caption*{\footnotesize RGB~3}
\end{subfigure}\hfill
\begin{subfigure}[b]{0.24\textwidth}
    \includegraphics[width=\linewidth]{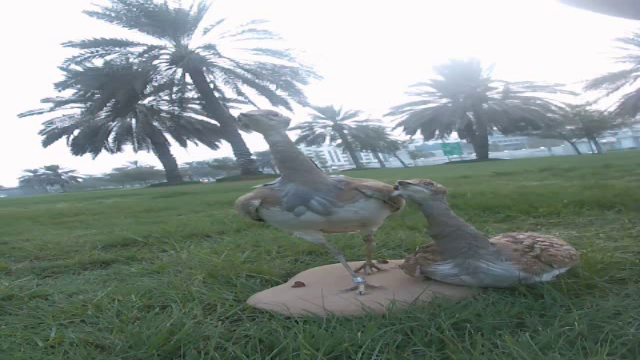}
    \caption*{\footnotesize RGB~4}
\end{subfigure}

\vspace{2mm}

\begin{subfigure}[b]{0.24\textwidth}
    \includegraphics[width=\linewidth]{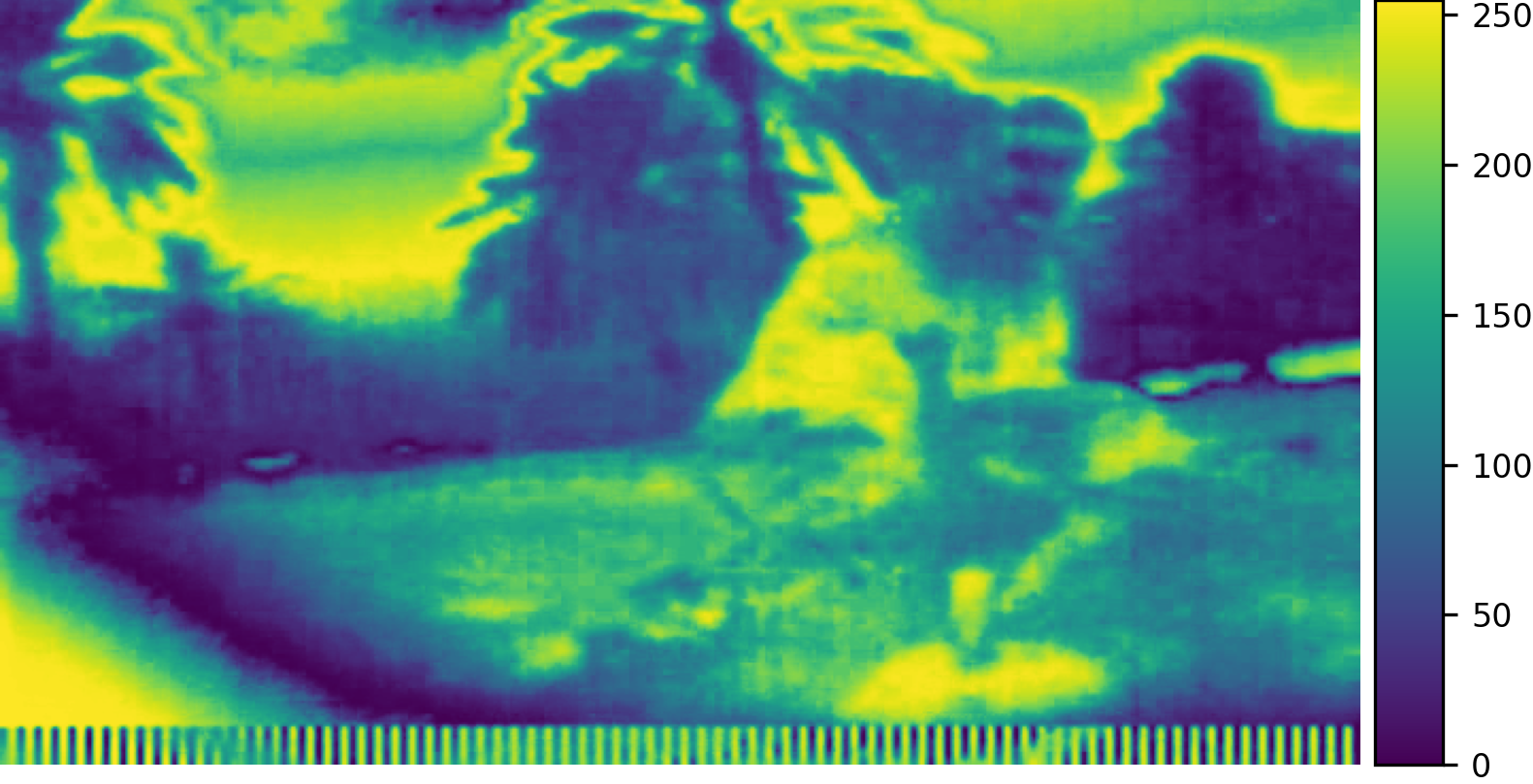}
    \caption*{\footnotesize Thermal~1}
\end{subfigure}\hfill
\begin{subfigure}[b]{0.24\textwidth}
    \includegraphics[width=\linewidth]{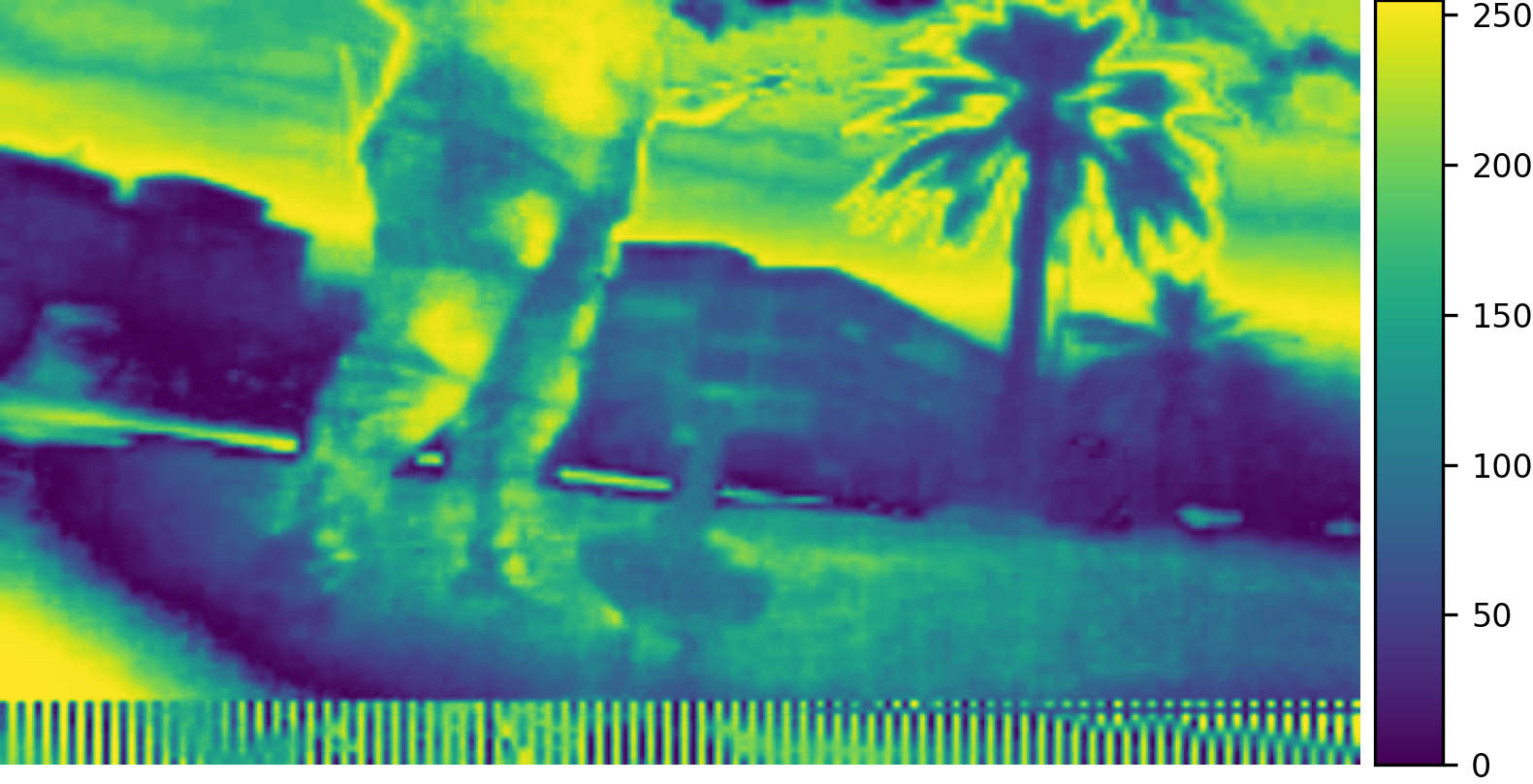}
    \caption*{\footnotesize Thermal~2}
\end{subfigure}\hfill
\begin{subfigure}[b]{0.24\textwidth}
    \includegraphics[width=\linewidth]{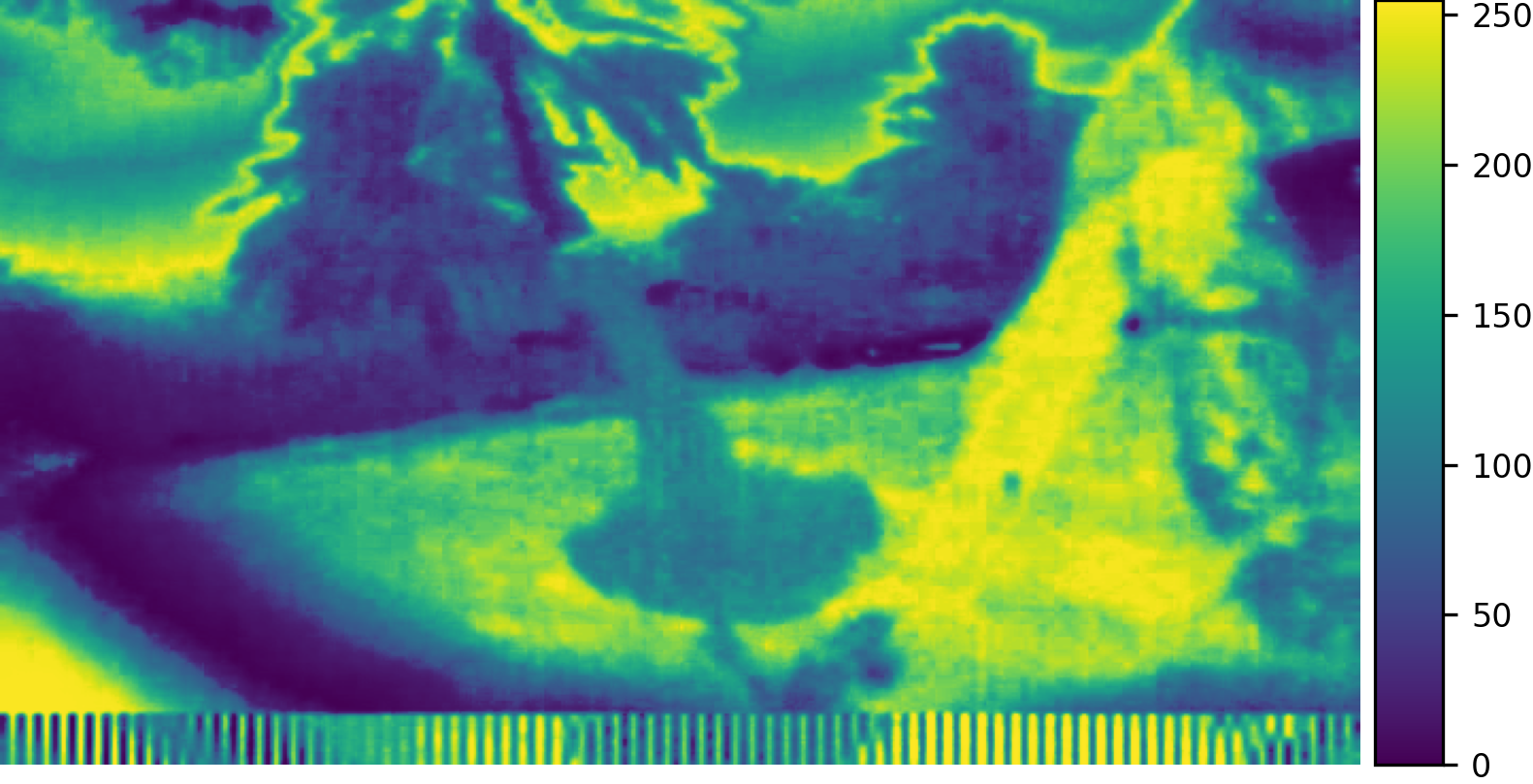}
    \caption*{\footnotesize Thermal~3}
\end{subfigure}\hfill
\begin{subfigure}[b]{0.24\textwidth}
    \includegraphics[width=\linewidth]{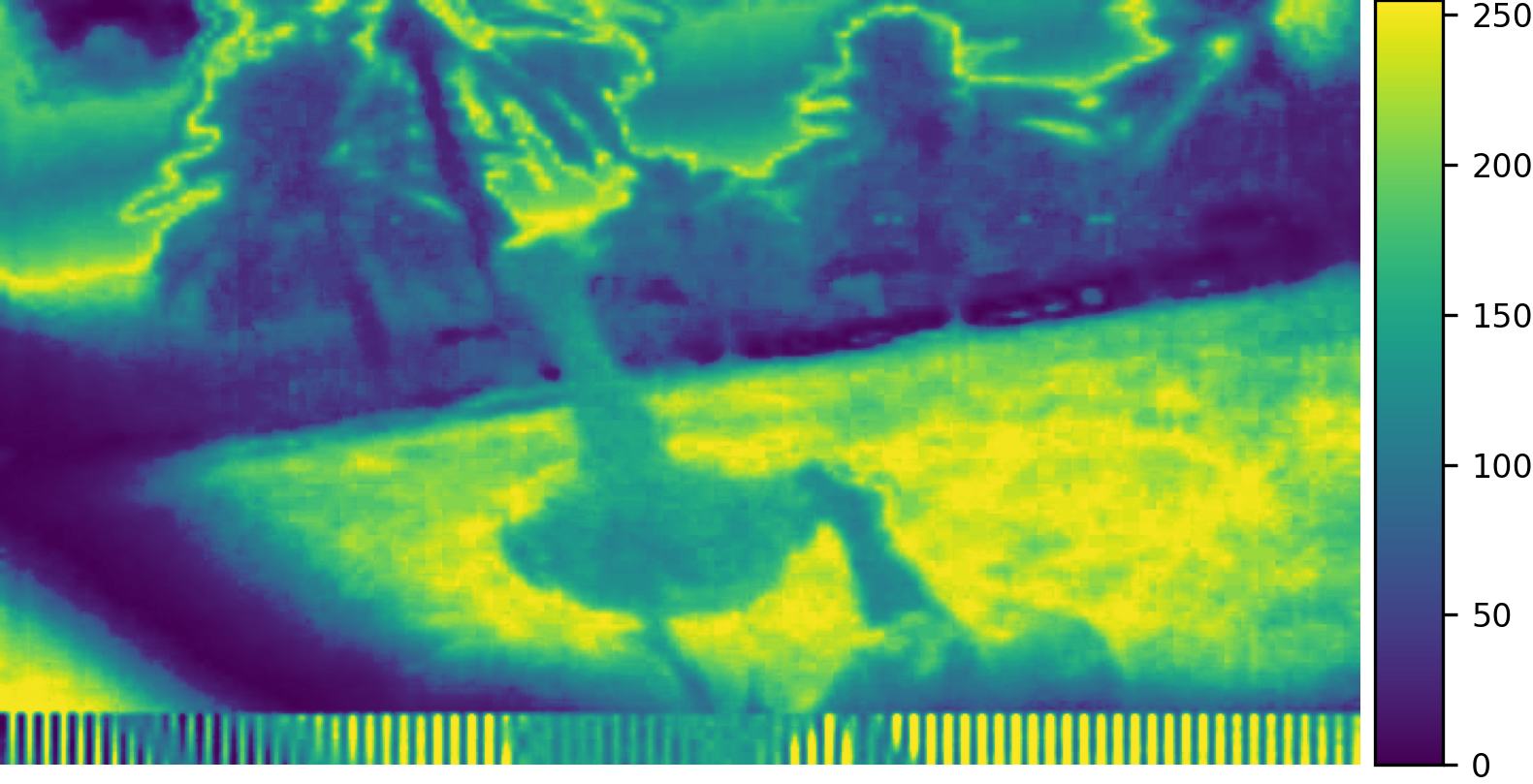}
    \caption*{\footnotesize Thermal~4}
\end{subfigure}

\vspace{2mm}

\begin{subfigure}[b]{0.23\textwidth}
    \includegraphics[width=\linewidth]{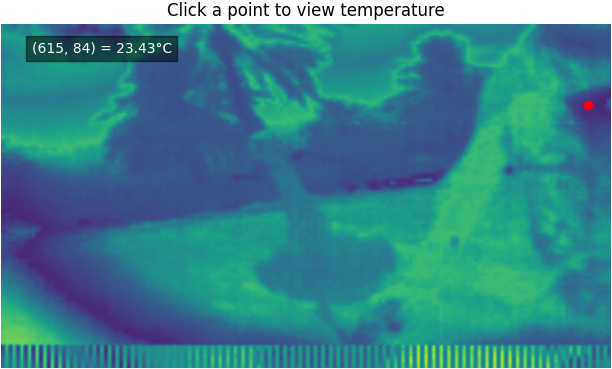}
    \caption*{\scriptsize Query~1}
\end{subfigure}\hfill
\begin{subfigure}[b]{0.23\textwidth}
    \includegraphics[width=\linewidth]{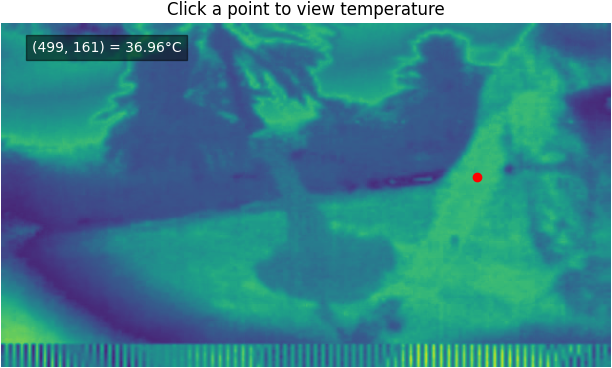}
    \caption*{\scriptsize Query~2}
\end{subfigure}\hfill
\begin{subfigure}[b]{0.23\textwidth}
    \includegraphics[width=\linewidth]{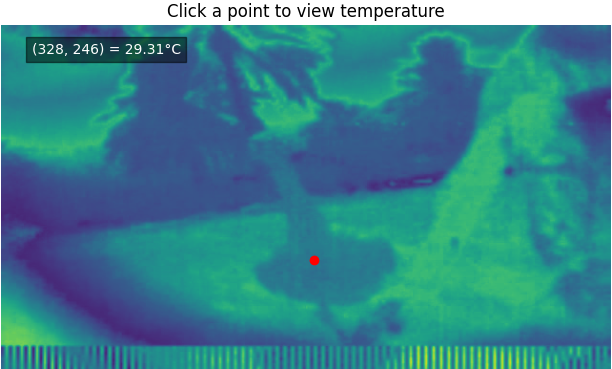}
    \caption*{\scriptsize Query~3}
\end{subfigure}\hfill
\begin{subfigure}[b]{0.23\textwidth}
    \includegraphics[width=\linewidth]{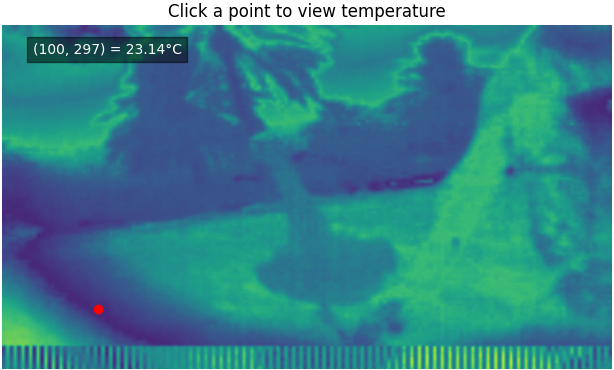}
    \caption*{\scriptsize Query~4}
\end{subfigure}

\caption{Thermal perception workflow. Top: synchronized RGB and thermally corrected frames demonstrating spatial alignment and heat distribution. Bottom: GUI snapshots enabling pixel-level temperature queries (red marker).}
\label{fig:thermal_alignment_queries}
\end{figure*}

\textbf{Thermal sensing and multi-modal perception.}
Thermal imaging was integrated to enable robust monitoring under variable illumination and at night. The perception pipeline combines an RGB camera with a FLIR Lepton long-wave infrared (LWIR) module and includes radiometric readout for temperature queries. Figure~\ref{fig:thermal_alignment_queries} illustrates the workflow: synchronized RGB and thermally corrected frames confirm spatial alignment and heat signatures, while an interactive GUI allows pixel-level temperature inspection directly from the thermal feed.

\begin{figure*}[t]
\centering
\begin{subfigure}[b]{0.47\textwidth}
    \includegraphics[width=\linewidth]{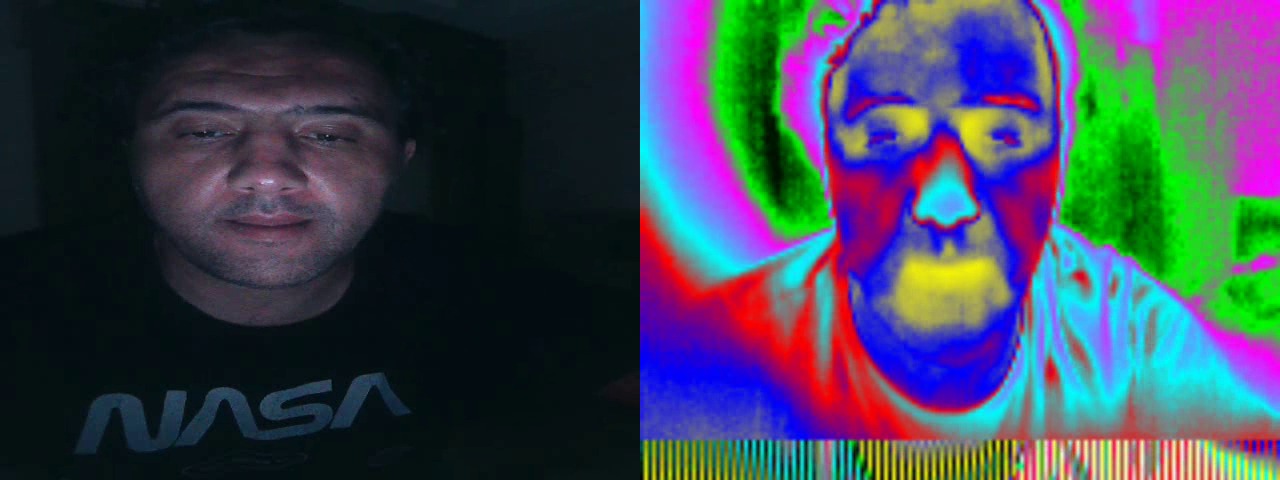}
    \caption*{\footnotesize Scene~1: Light ON}
\end{subfigure}\hfill
\begin{subfigure}[b]{0.47\textwidth}
    \includegraphics[width=\linewidth]{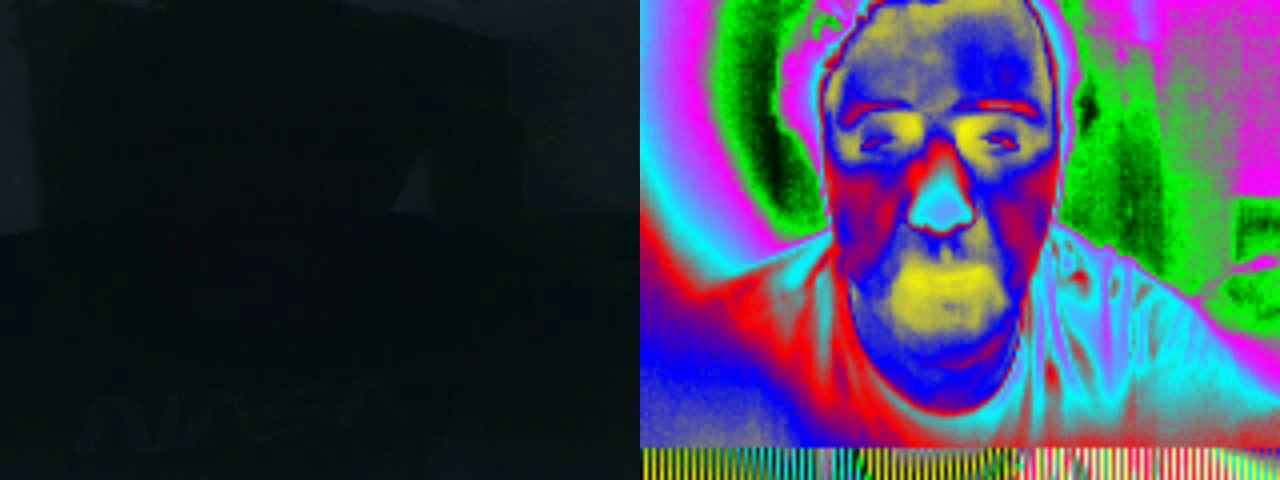}
    \caption*{\footnotesize Scene~1: Light OFF}
\end{subfigure}

\begin{subfigure}[b]{0.47\textwidth}
    \includegraphics[width=\linewidth]{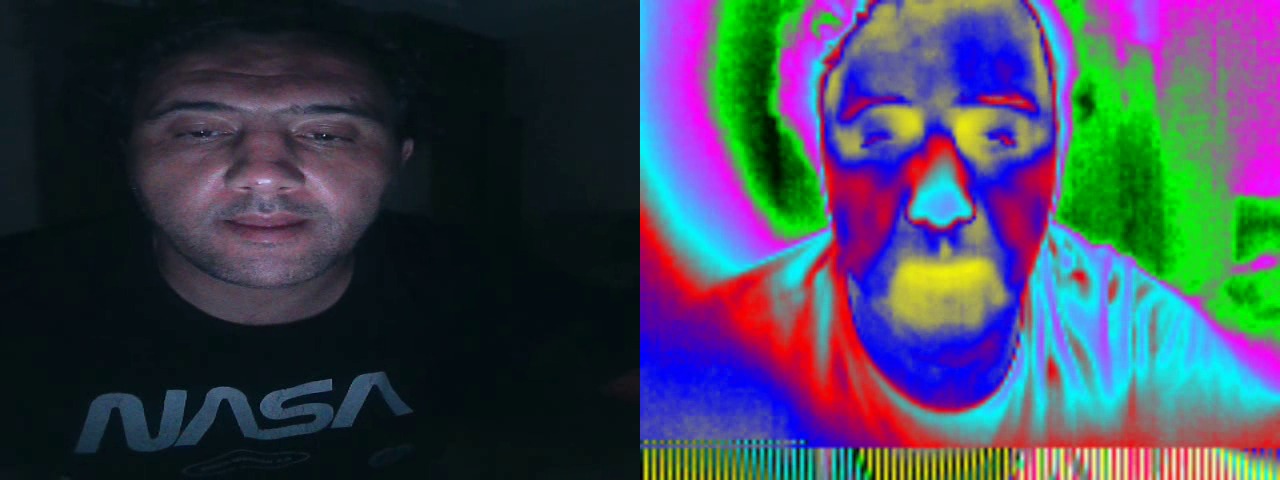}
    \caption*{\footnotesize Scene~2: Light ON}
\end{subfigure}\hfill
\begin{subfigure}[b]{0.47\textwidth}
    \includegraphics[width=\linewidth]{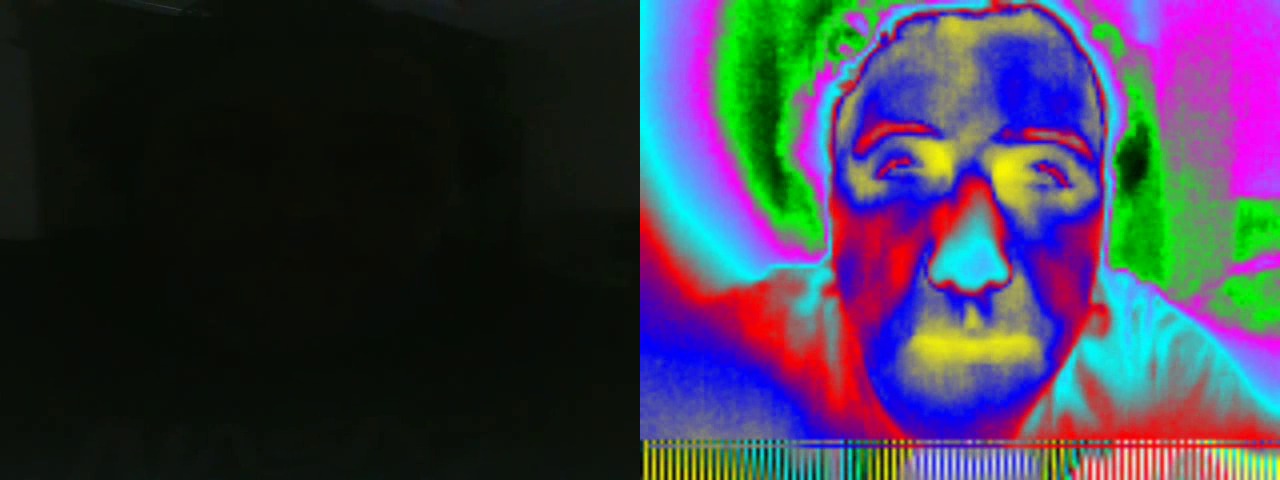}
    \caption*{\footnotesize Scene~2: Light OFF}
\end{subfigure}

\begin{subfigure}[b]{0.47\textwidth}
    \includegraphics[width=\linewidth]{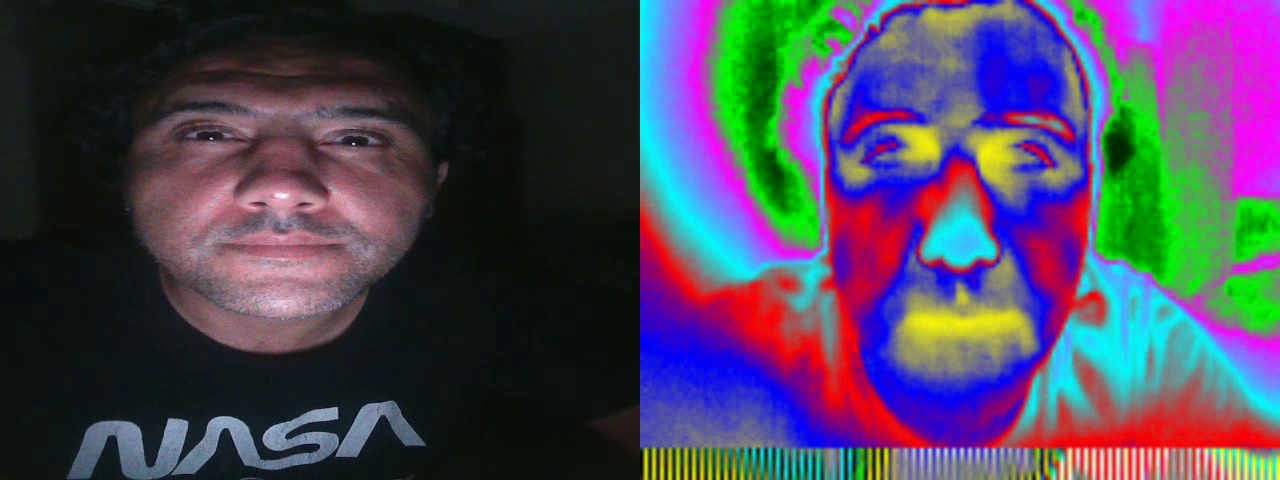}
    \caption*{\footnotesize Scene~3: Light ON}
\end{subfigure}\hfill
\begin{subfigure}[b]{0.47\textwidth}
    \includegraphics[width=\linewidth]{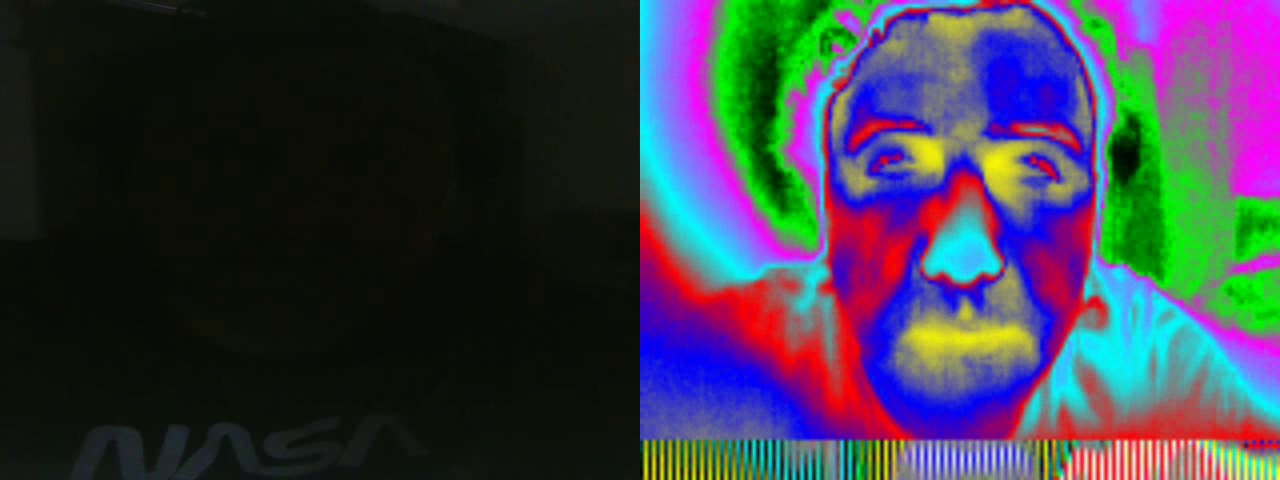}
    \caption*{\footnotesize Scene~3: Light OFF}
\end{subfigure}

\begin{subfigure}[b]{0.47\textwidth}
    \includegraphics[width=\linewidth]{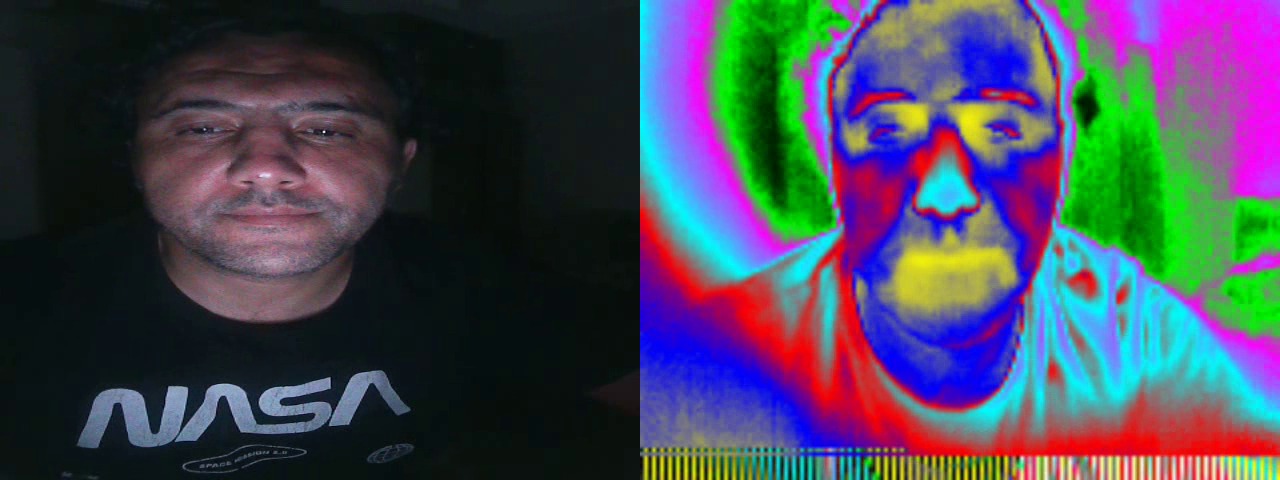}
    \caption*{\footnotesize Scene~4: Light ON}
\end{subfigure}\hfill
\begin{subfigure}[b]{0.47\textwidth}
    \includegraphics[width=\linewidth]{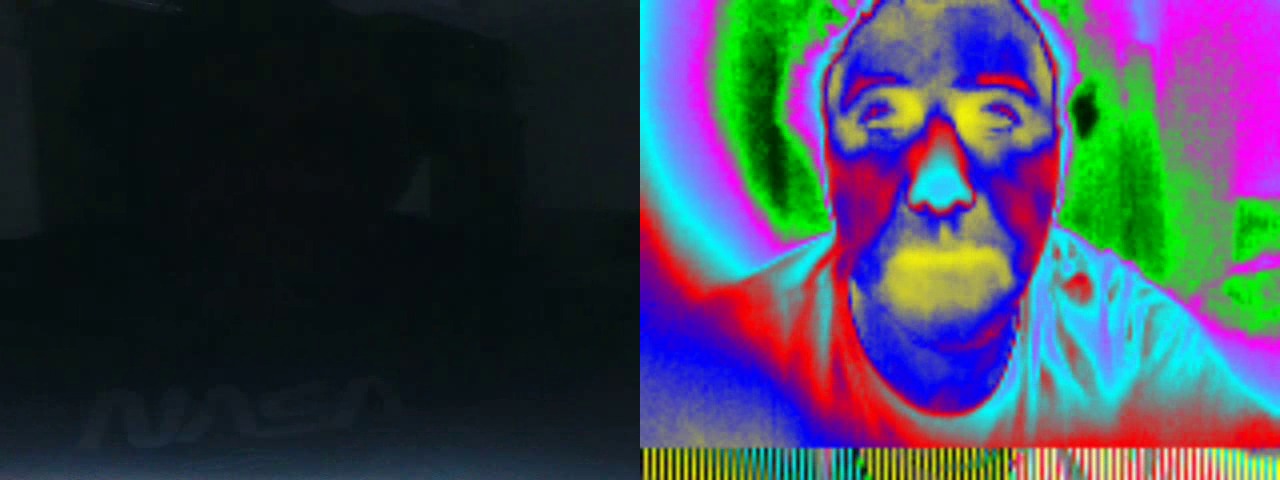}
    \caption*{\footnotesize Scene~4: Light OFF}
\end{subfigure}

\begin{subfigure}[b]{0.47\textwidth}
    \includegraphics[width=\linewidth]{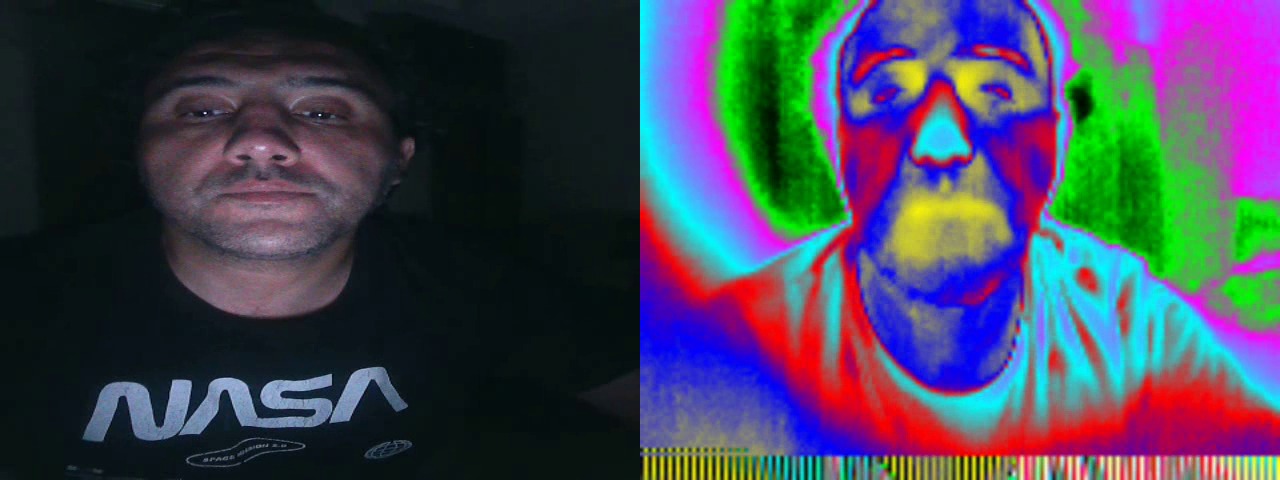}
    \caption*{\footnotesize Scene~5: Light ON}
\end{subfigure}\hfill
\begin{subfigure}[b]{0.47\textwidth}
    \includegraphics[width=\linewidth]{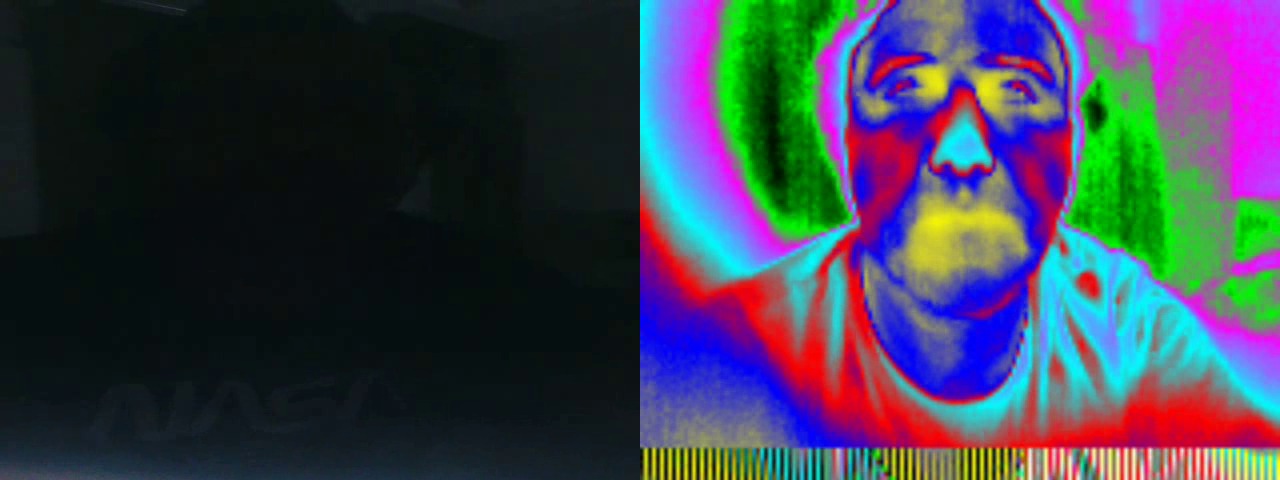}
    \caption*{\footnotesize Scene~5: Light OFF}
\end{subfigure}

\caption{Lighting-invariance validation of the thermal module. Each scene shows a merged RGB (left) and thermal (right) view captured at night with a phone flashlight turned on (left) and the same scene in complete darkness (right). Visible illumination strongly changes RGB appearance but leaves the long-wave infrared signal unaffected, confirming robustness of thermal perception for nocturnal monitoring.}
\label{fig:thermal_lighting_invariance}
\end{figure*}

To test robustness against visible illumination, we captured paired night scenes with a phone flashlight \emph{on} and \emph{off}, recording both RGB and thermal streams. As shown in Figure~\ref{fig:thermal_lighting_invariance}, visible light strongly alters RGB appearance but has negligible influence on the thermal signal, consistent with LWIR imaging (8–14~$\mu$m) being largely insensitive to visible wavelengths. These results support reliable day–night operation for behavioral monitoring; species-specific temperature calibration for live birds remains an avenue for future work.

\subsection{Thermal–Visible Fusion Results}

\begin{figure*}[t]
\centering

\begin{subfigure}[b]{0.24\textwidth}
  \includegraphics[width=\linewidth]{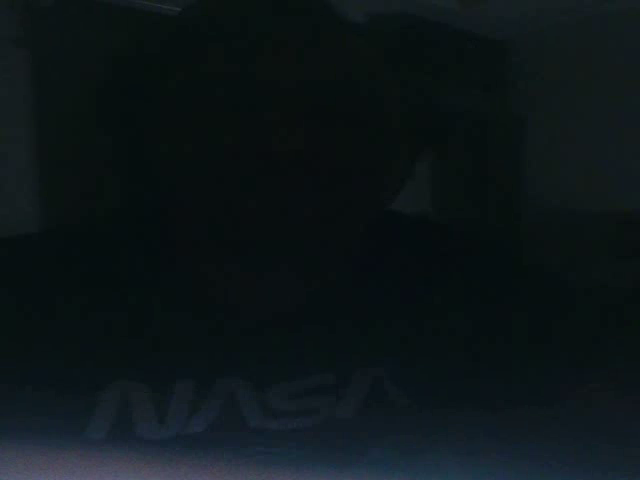}
  \caption*{\footnotesize Orig \#25}
\end{subfigure}\hfill
\begin{subfigure}[b]{0.24\textwidth}
  \includegraphics[width=\linewidth]{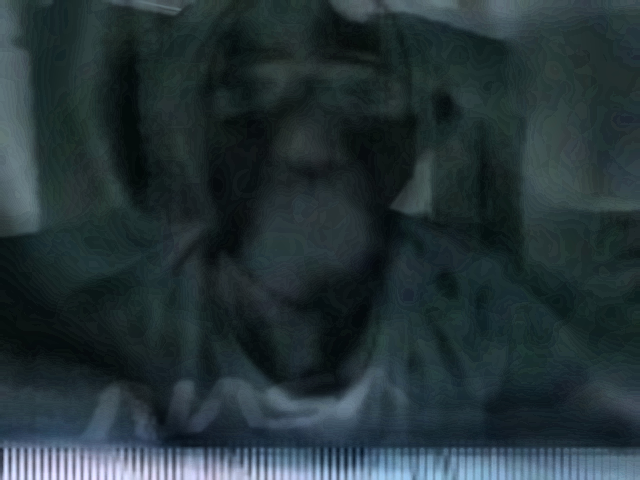}
  \caption*{\footnotesize Fused \#25}
\end{subfigure}\hfill
\begin{subfigure}[b]{0.24\textwidth}
  \includegraphics[width=\linewidth]{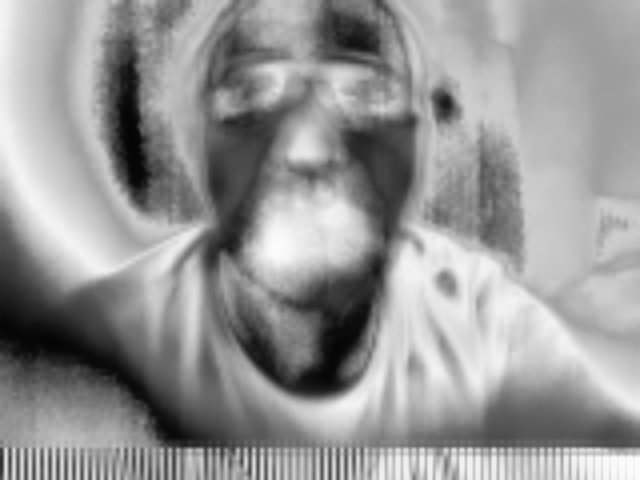}
  \caption*{\footnotesize $\hat{L}$ (gray) \#25}
\end{subfigure}\hfill
\begin{subfigure}[b]{0.24\textwidth}
  \includegraphics[width=\linewidth]{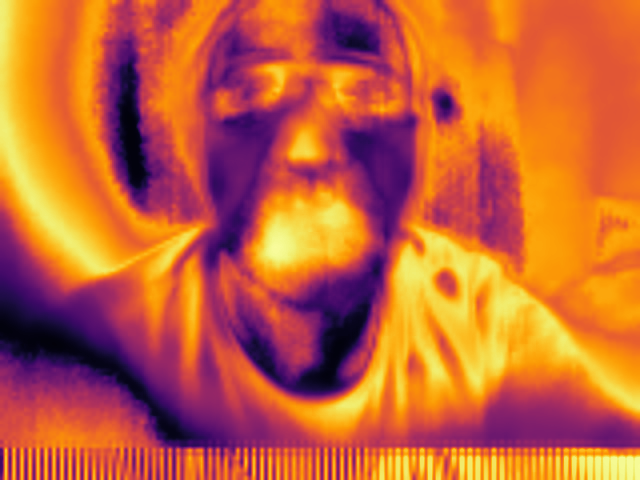}
  \caption*{\footnotesize $\hat{L}$ (Inferno) \#25}
\end{subfigure}

\vspace{2mm}

\begin{subfigure}[b]{0.24\textwidth}
  \includegraphics[width=\linewidth]{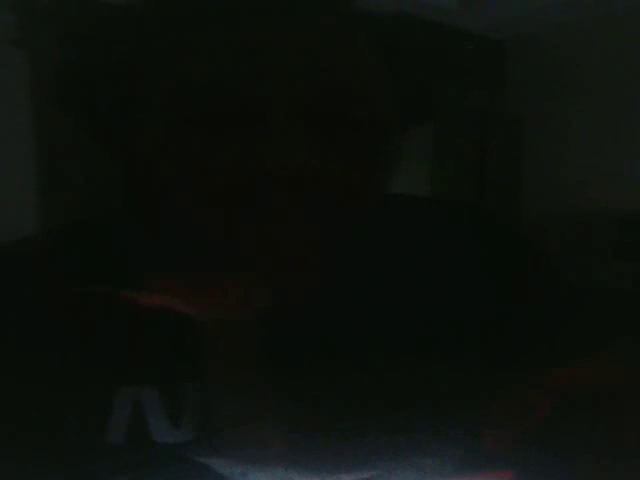}
  \caption*{\footnotesize Orig \#632}
\end{subfigure}\hfill
\begin{subfigure}[b]{0.24\textwidth}
  \includegraphics[width=\linewidth]{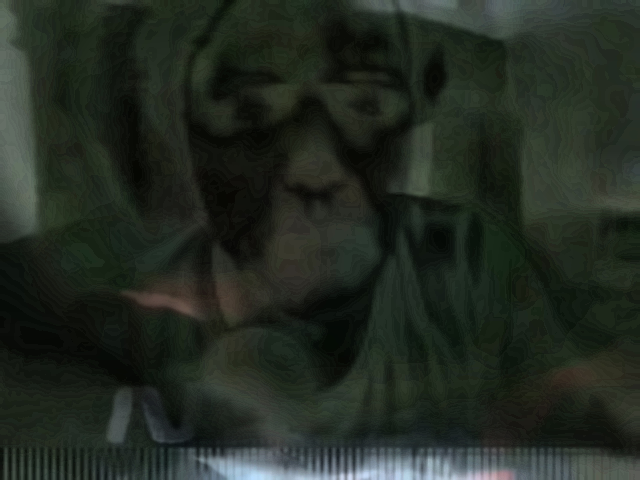}
  \caption*{\footnotesize Fused \#632}
\end{subfigure}\hfill
\begin{subfigure}[b]{0.24\textwidth}
  \includegraphics[width=\linewidth]{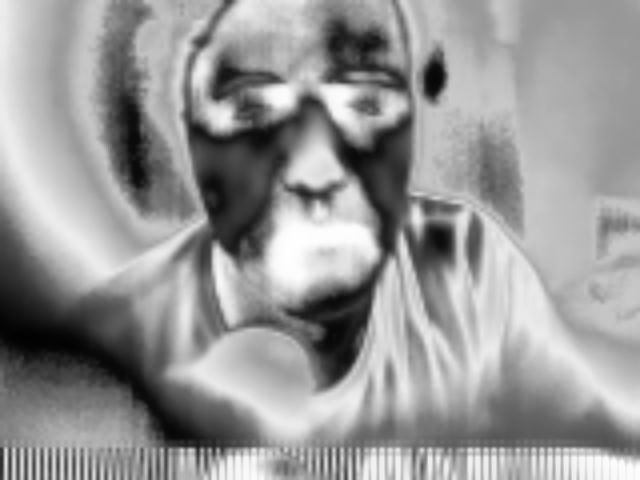}
  \caption*{\footnotesize $\hat{L}$ (gray) \#632}
\end{subfigure}\hfill
\begin{subfigure}[b]{0.24\textwidth}
  \includegraphics[width=\linewidth]{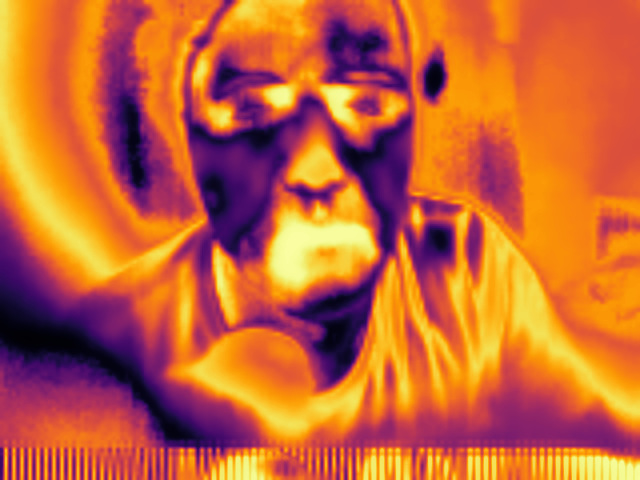}
  \caption*{\footnotesize $\hat{L}$ (Inferno) \#632}
\end{subfigure}

\vspace{2mm}

\begin{subfigure}[b]{0.24\textwidth}
  \includegraphics[width=\linewidth]{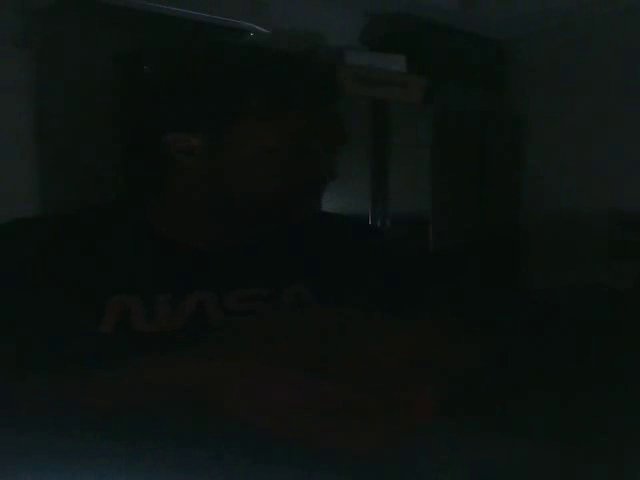}
  \caption*{\footnotesize Orig \#723}
\end{subfigure}\hfill
\begin{subfigure}[b]{0.24\textwidth}
  \includegraphics[width=\linewidth]{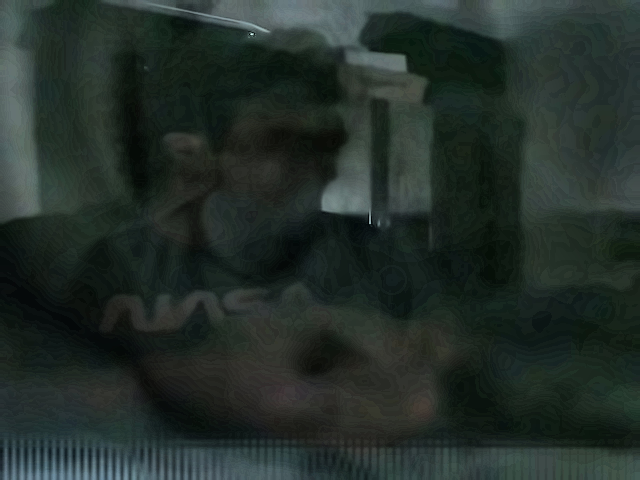}
  \caption*{\footnotesize Fused \#723}
\end{subfigure}\hfill
\begin{subfigure}[b]{0.24\textwidth}
  \includegraphics[width=\linewidth]{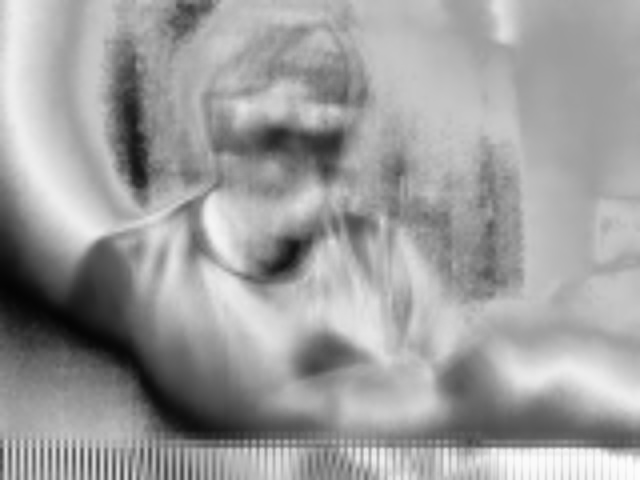}
  \caption*{\footnotesize $\hat{L}$ (gray) \#723}
\end{subfigure}\hfill
\begin{subfigure}[b]{0.24\textwidth}
  \includegraphics[width=\linewidth]{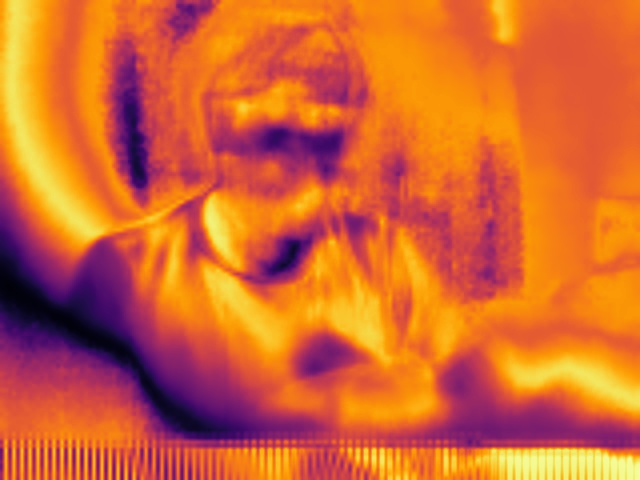}
  \caption*{\footnotesize $\hat{L}$ (Inferno) \#723}
\end{subfigure}

\vspace{2mm}

\begin{subfigure}[b]{0.24\textwidth}
  \includegraphics[width=\linewidth]{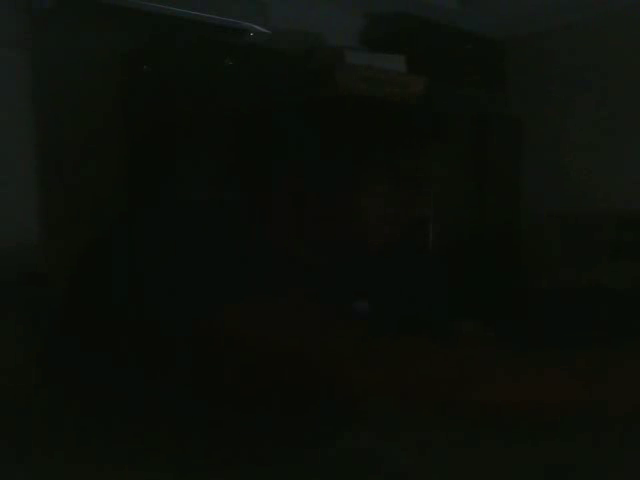}
  \caption*{\footnotesize Orig \#798}
\end{subfigure}\hfill
\begin{subfigure}[b]{0.24\textwidth}
  \includegraphics[width=\linewidth]{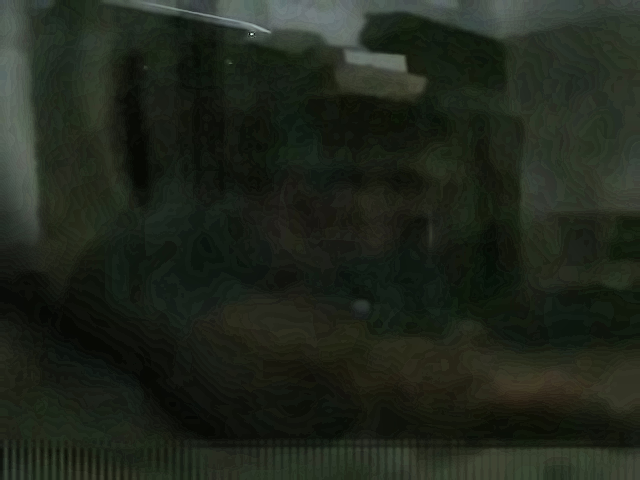}
  \caption*{\footnotesize Fused \#798}
\end{subfigure}\hfill
\begin{subfigure}[b]{0.24\textwidth}
  \includegraphics[width=\linewidth]{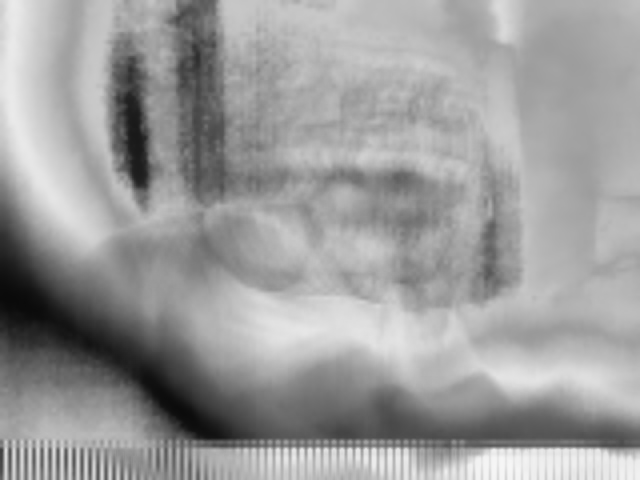}
  \caption*{\footnotesize $\hat{L}$ (gray) \#798}
\end{subfigure}\hfill
\begin{subfigure}[b]{0.24\textwidth}
  \includegraphics[width=\linewidth]{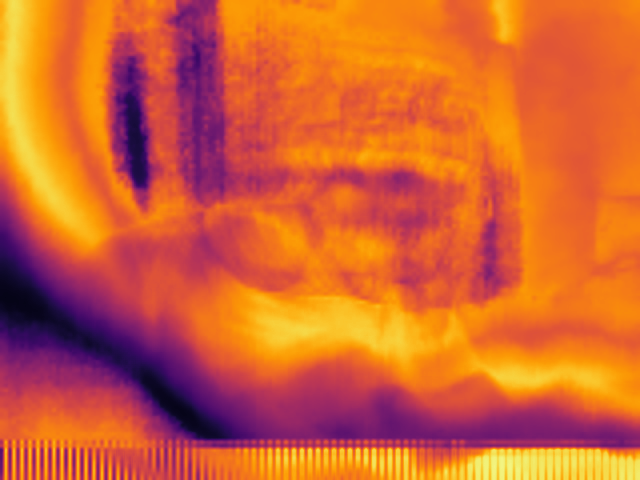}
  \caption*{\footnotesize $\hat{L}$ (Inferno) \#798}
\end{subfigure}

\caption{Thermal–visible fusion at night. For each selected index (rows), we show the original RGB, the fused output, and the illumination map $\hat{L}$ in grayscale and Inferno. The fused view preserves structural detail while emphasizing thermal saliency; $\hat{L}$ provides an interpretable heat-driven gain field.}
\label{fig:fusion_grid}
\end{figure*}

Figure~\ref{fig:fusion_grid} presents four synchronized views for each representative night frame: the original RGB input, the fused output, and the illumination map $\hat{L}$ in both grayscale and an Inferno colormap. Across the selected frames (\#25, \#632, \#723, \#798), NightFusion preserves visible structure while emphasizing thermally salient anatomy and nearby warm surfaces. The illumination map remains temporally stable and spatially aligned with heat sources, acting as an interpretable gain field that drives the enhancement. To ensure fair visual comparison, all $\hat{L}$ panels are rendered with a fixed display range.

Qualitatively, frame~\#25 reveals subject contours emerging from a dim background; $\hat{L}$ concentrates on facial and torso regions, and the fused output strengthens those areas without washing out texture. In frame~\#632, the thermal contrast is stronger around the head and upper body, and the fusion retains local detail without saturation. Frame~\#723 shows an RGB view that is nearly black; the fusion reconstructs recognizable geometry guided by a smooth $\hat{L}$ that avoids flicker. Frame~\#798 contains more clutter and low illumination, yet $\hat{L}$ remains stable and the fused output avoids amplifying noise. Together, these examples illustrate NightFusion’s key advantages: lighting invariance in dark scenes, an interpretable heat-driven enhancement mechanism, and real-time operation on embedded hardware. At the same time, they highlight practical limits such as residual misalignment from RGB–LWIR parallax at short range, dependence on emissivity and radiometric calibration for absolute temperature accuracy, and the usual trade-off between temporal smoothing and responsiveness to fast thermal changes. Additional sequences and interactive per-pixel temperature readouts are provided in the supplementary video.

\paragraph{Comparison to existing fusion approaches.}
A formal quantitative benchmark against other RGB–thermal fusion techniques (e.g., guided filtering, multi-scale weighted least squares, or deep fusion networks) was not performed in this study, as our primary goal was to achieve real-time, field-deployable perception on the embedded NVIDIA Jetson platform. NightFusion was therefore designed to be lightweight and interpretable: it applies temporally smoothed illumination estimation, thermal-guided gain modulation, and contrast-limited histogram equalization to enhance salient heat signatures while preserving visible structural cues. These design choices balance visual clarity and computational cost, achieving frame rates of 15–22~FPS with sub-100~ms latency. Such performance would be difficult for many existing deep fusion models to maintain in outdoor, battery-powered conditions. Future work will include systematic quantitative evaluation of NightFusion against established fusion baselines using metrics such as PSNR, SSIM, and perceptual quality indices as soon as standardized outdoor RGB–thermal datasets become available.

\subsection{Physical interpretation.}
The FLIR Lepton detects long-wave infrared radiation (8–14~$\mu$m), determined by surface temperature and emissivity rather than reflected visible light. Phone flashlights emit primarily in the 0.4–1~$\mu$m range and therefore do not directly influence LWIR measurements. Only secondary effects such as slight heating of nearby surfaces or reflections from warm objects can alter thermal output. This explains the stability observed between the “light on” and “light off” conditions in Figure~\ref{fig:thermal_lighting_invariance}.

\begin{figure*}[t]
\centering
\begin{subfigure}[b]{0.22\textwidth}
    \includegraphics[width=\textwidth]{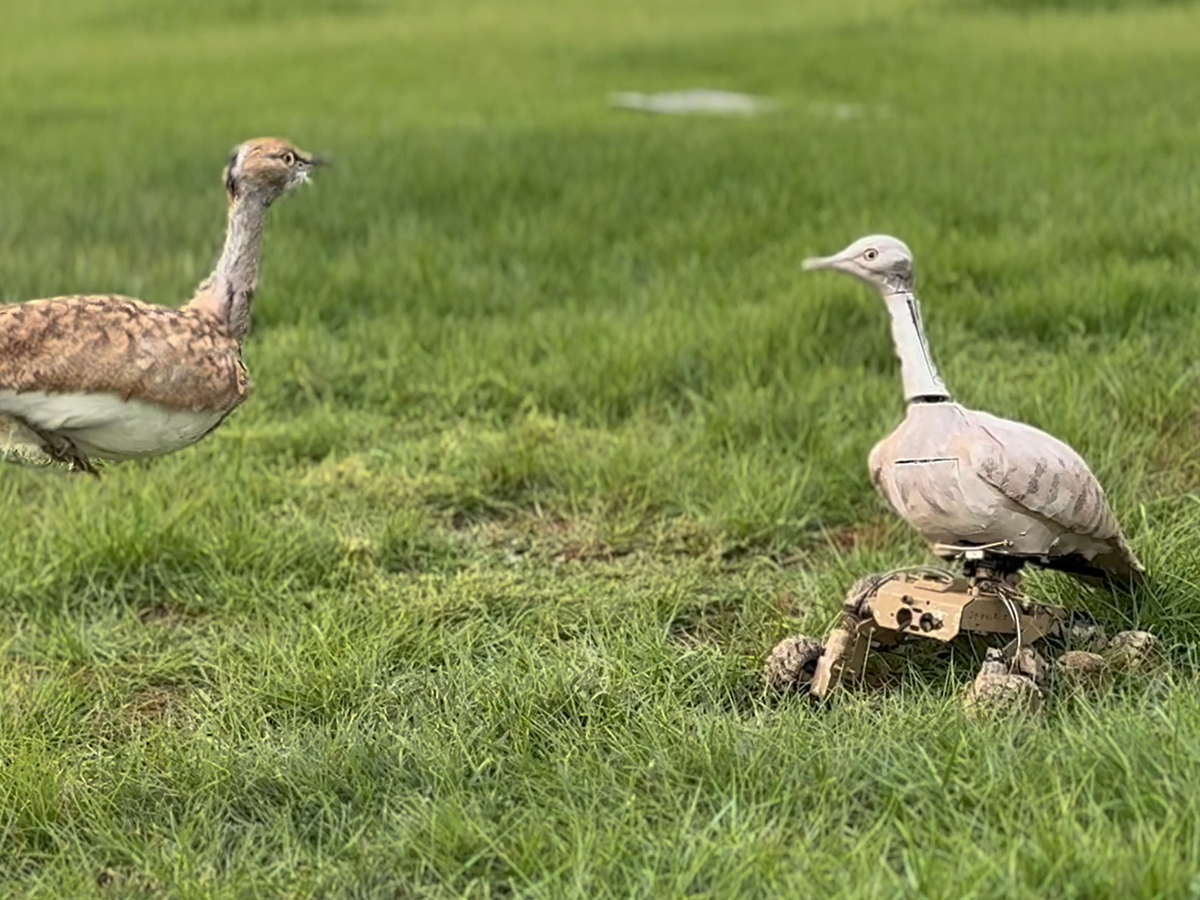}
    \caption*{External~1}
\end{subfigure}\hfill
\begin{subfigure}[b]{0.22\textwidth}
    \includegraphics[width=\textwidth]{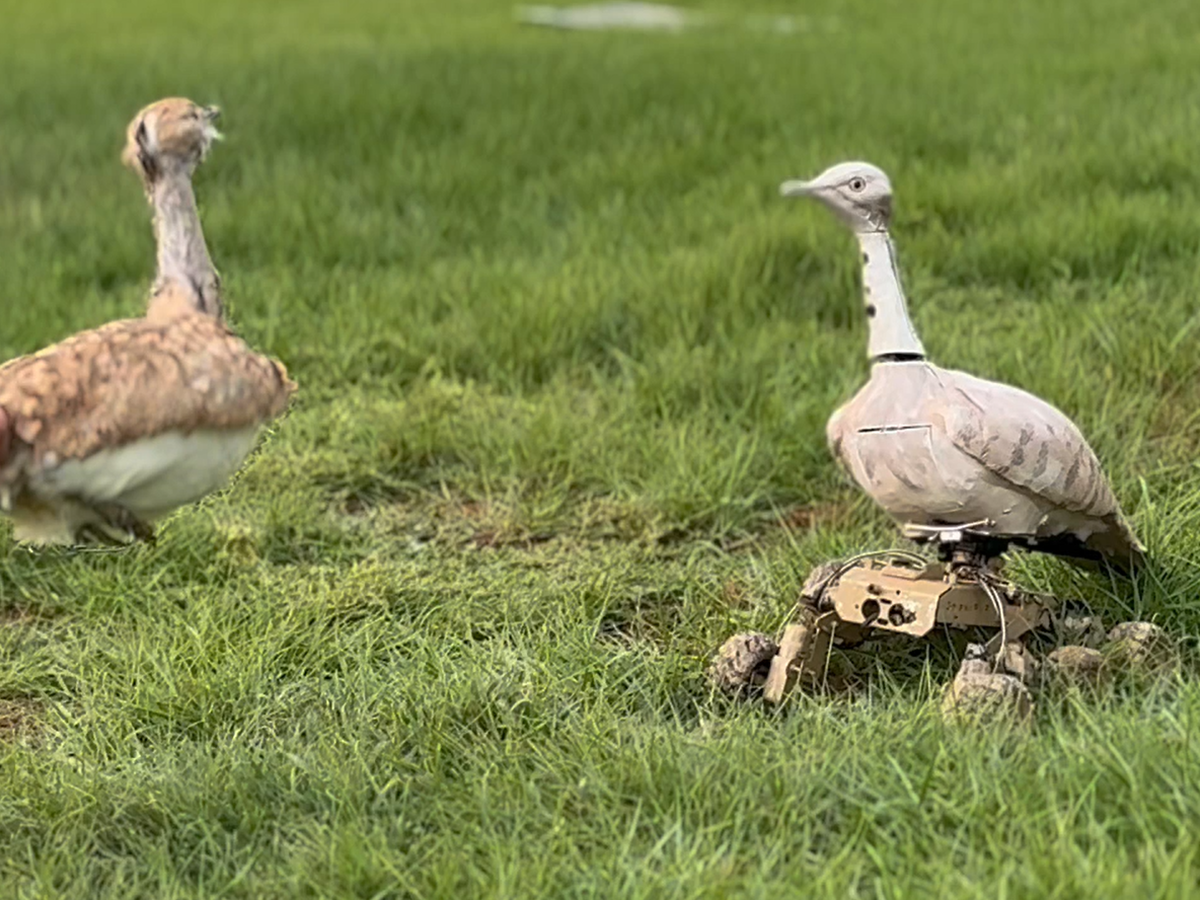}
    \caption*{External~2}
\end{subfigure}\hfill
\begin{subfigure}[b]{0.22\textwidth}
    \includegraphics[width=\textwidth]{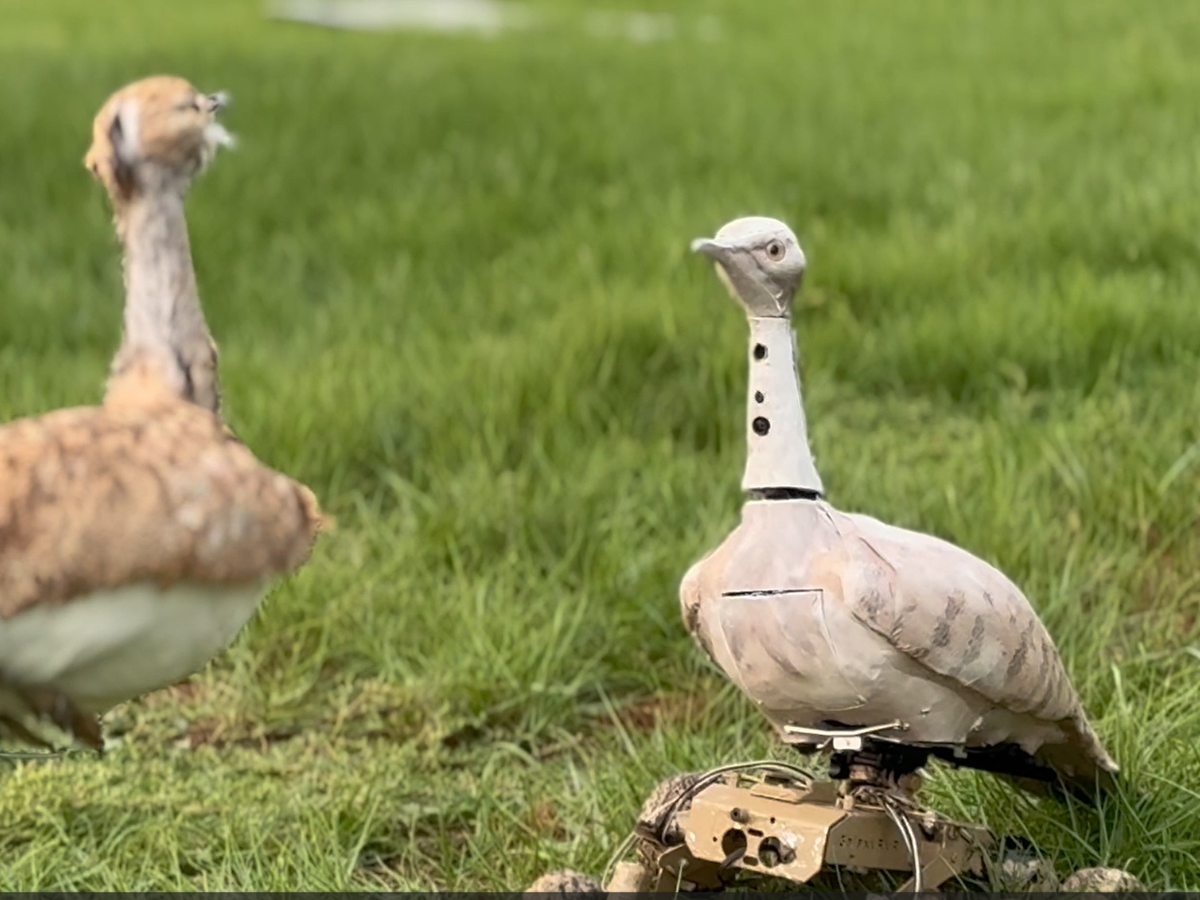}
    \caption*{External~3}
\end{subfigure}\hfill
\begin{subfigure}[b]{0.22\textwidth}
    \includegraphics[width=\textwidth]{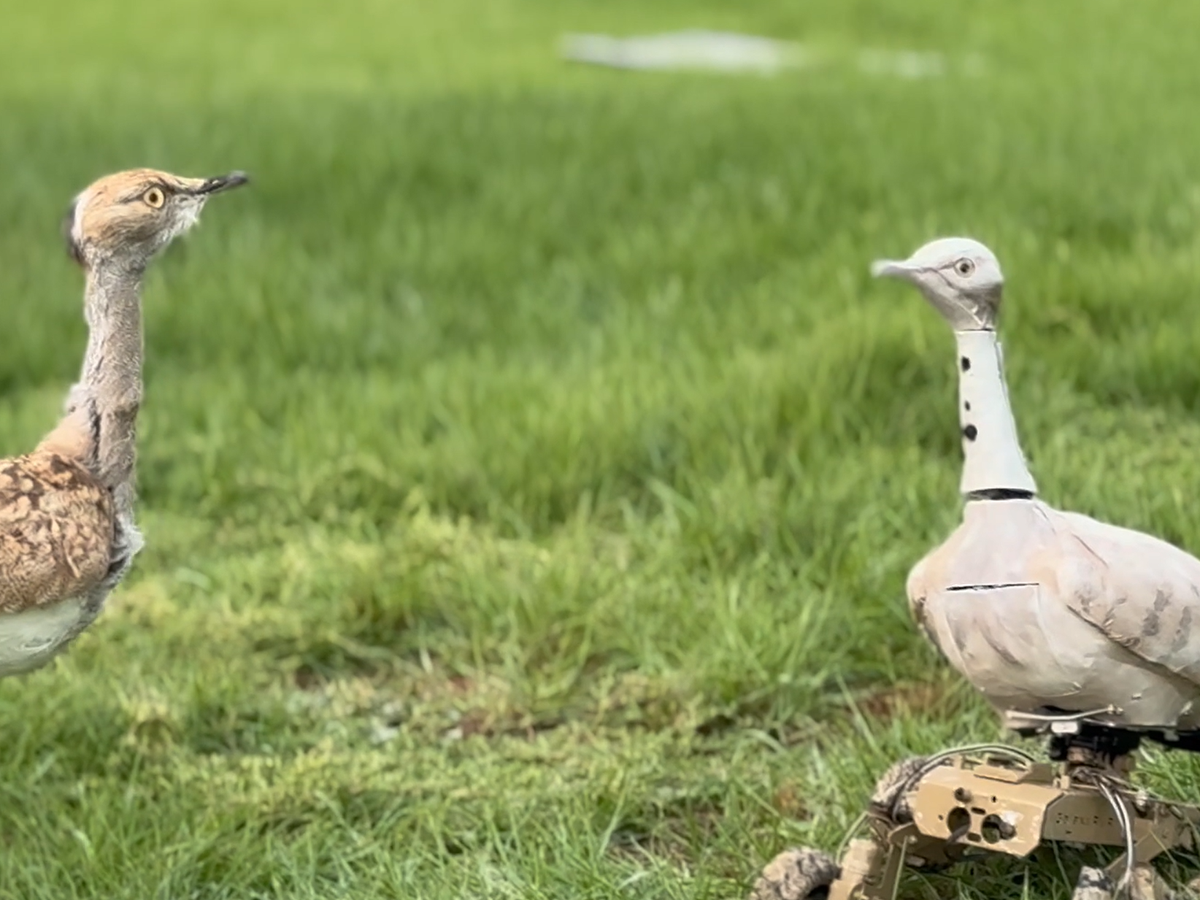}
    \caption*{External~4}
\end{subfigure}

\vspace{2mm}

\begin{subfigure}[b]{0.22\textwidth}
    \includegraphics[width=\textwidth]{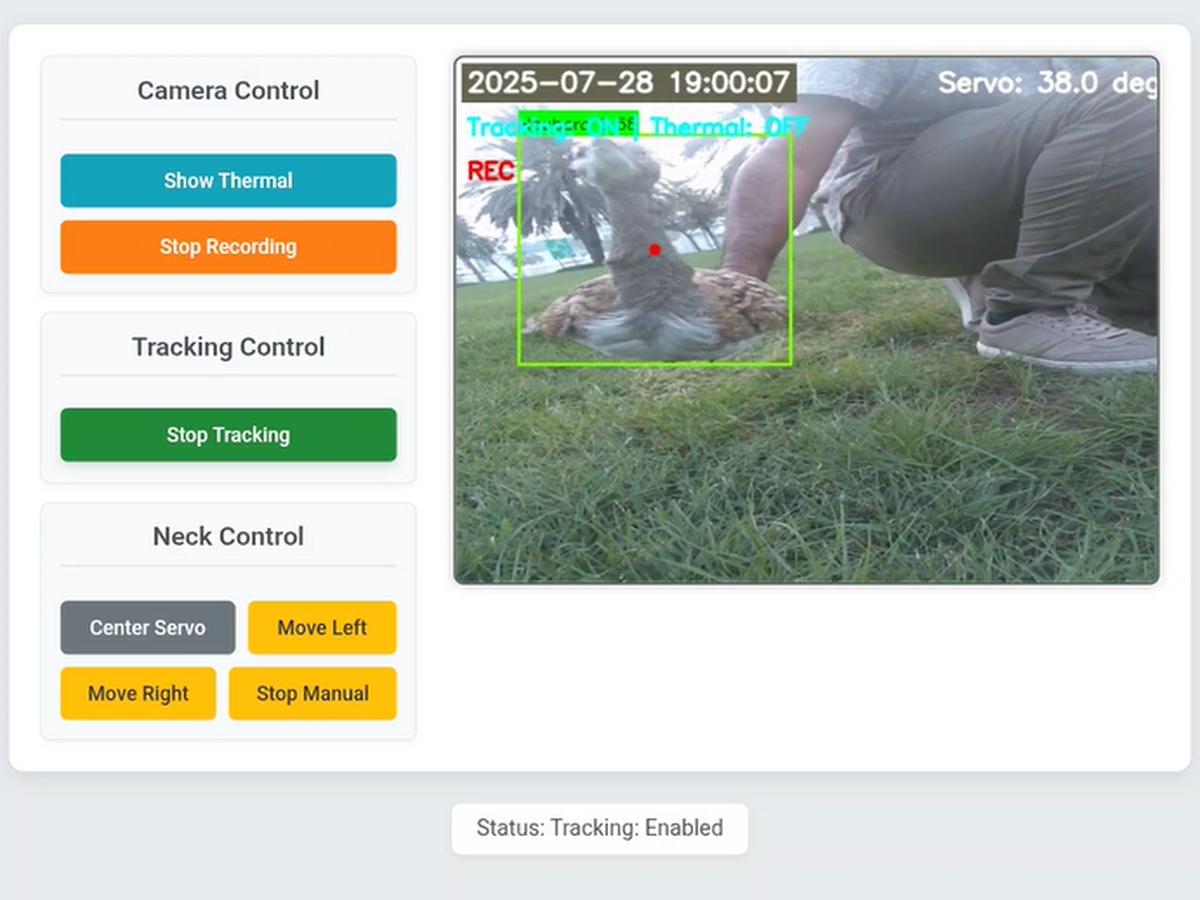}
    \caption*{GUI~1}
\end{subfigure}\hfill
\begin{subfigure}[b]{0.22\textwidth}
    \includegraphics[width=\textwidth]{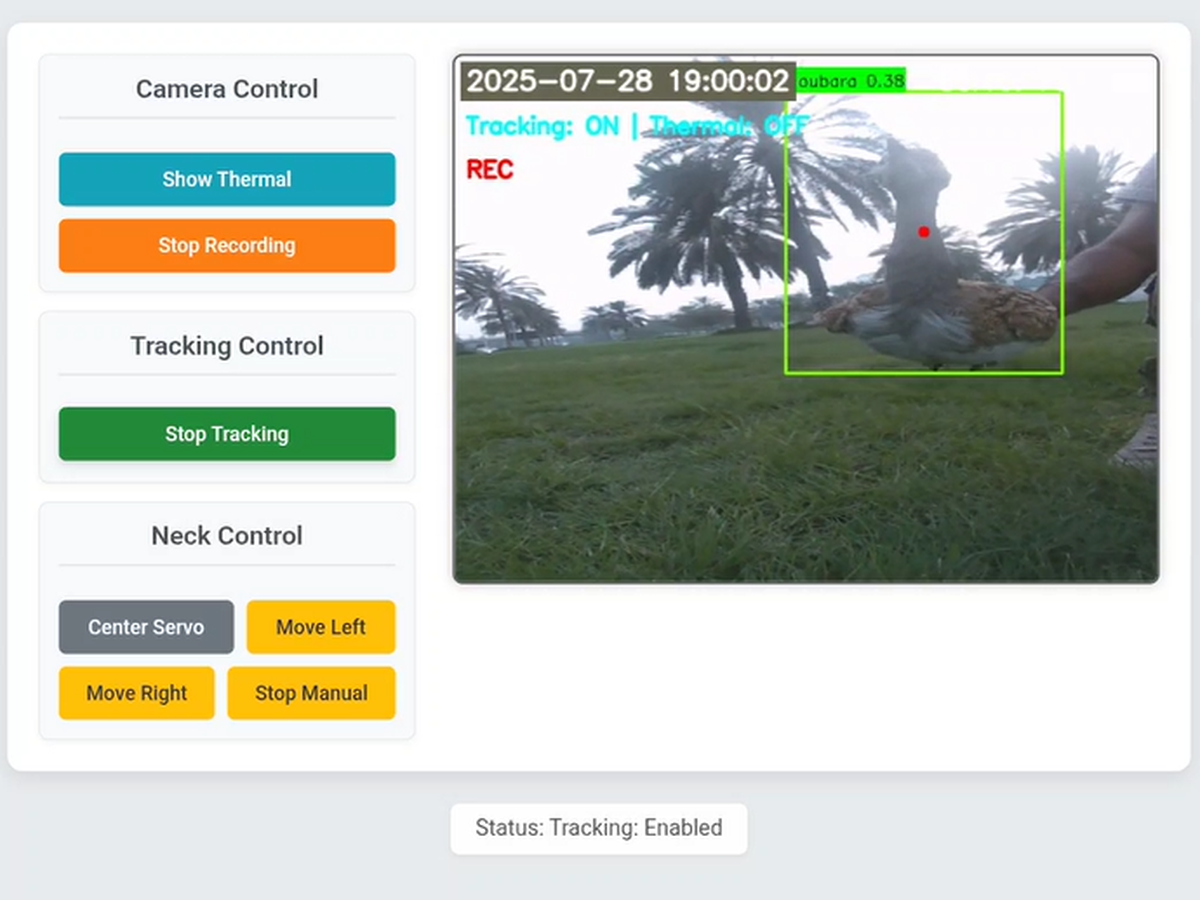}
    \caption*{GUI~2}
\end{subfigure}\hfill
\begin{subfigure}[b]{0.22\textwidth}
    \includegraphics[width=\textwidth]{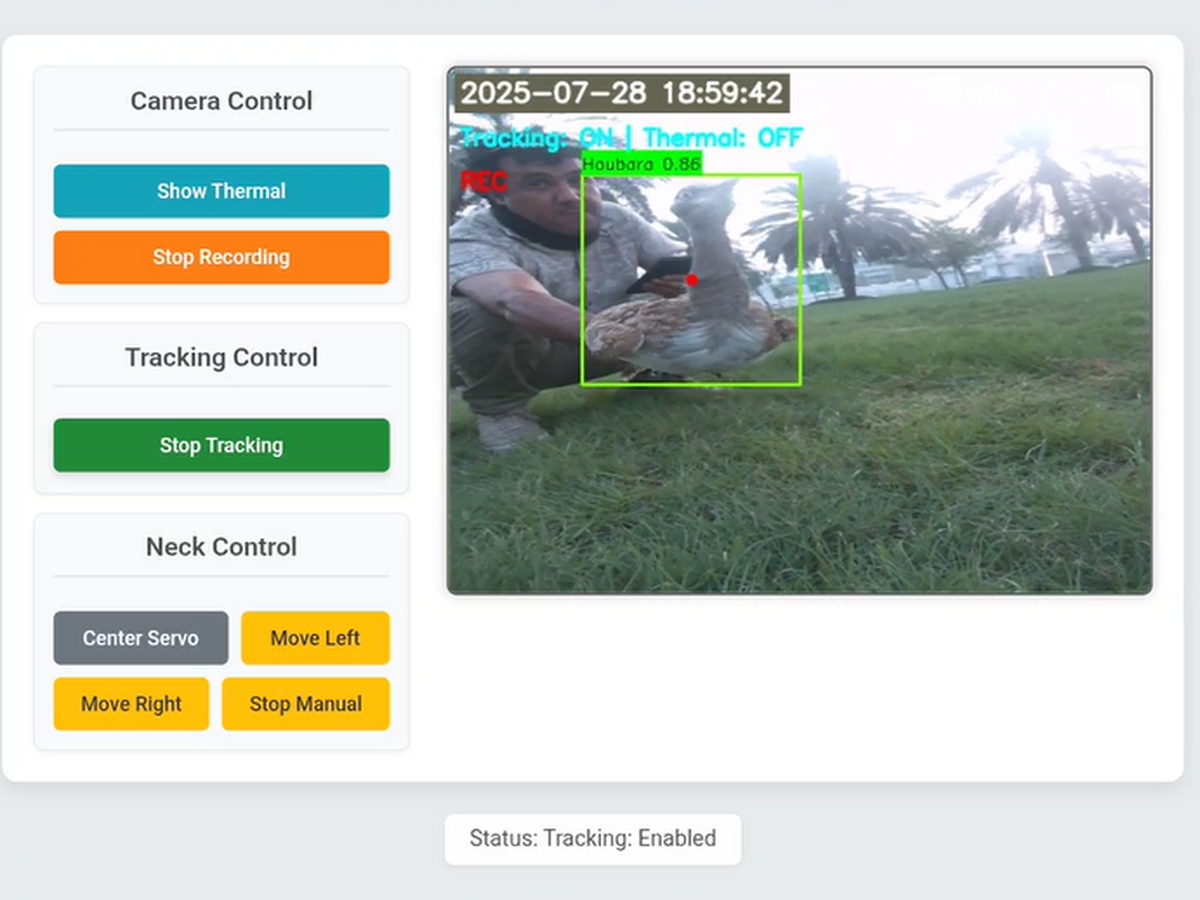}
    \caption*{GUI~3}
\end{subfigure}\hfill
\begin{subfigure}[b]{0.22\textwidth}
    \includegraphics[width=\textwidth]{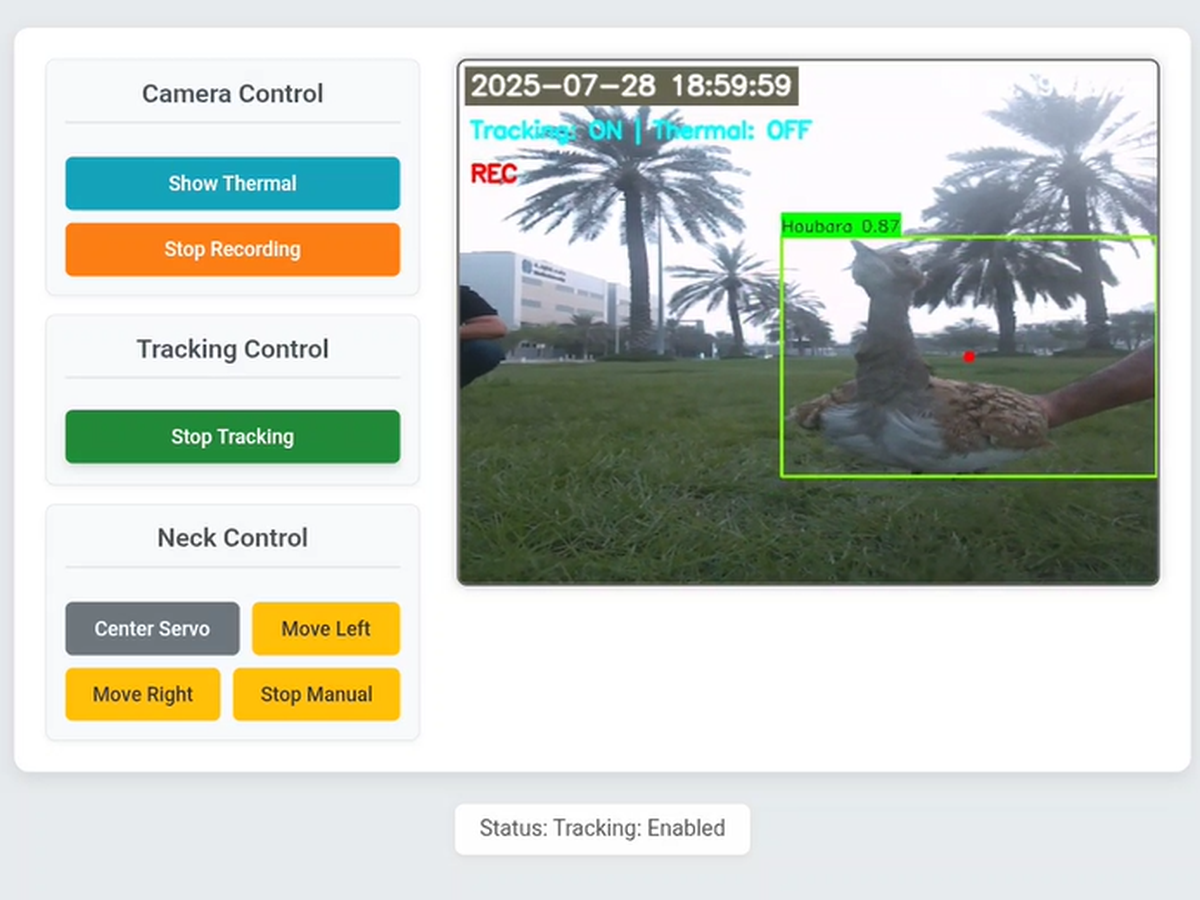}
    \caption*{GUI~4}
\end{subfigure}

\caption{Closed-loop visual tracking validation. Top: external views of the robot orienting toward a moving target. Bottom: onboard GUI with YOLOv11n detections driving the pan–tilt neck through PID control.}
\label{fig:movement_tracking_sequence}
\end{figure*}

\textbf{Vision-based tracking and closed-loop actuation.}
Closed-loop target following integrates YOLOv11n object detection with a PID-controlled pan–tilt neck. Figure~\ref{fig:movement_tracking_sequence} shows the robot orienting toward a moving target in outdoor tests: the top row captures external views of head movement, and the bottom row displays the onboard GUI where real-time detections drive the PID controller. The system maintained stable alignment even under variable illumination and partial occlusion, validating the real-time perception–action loop.

\section{Quantitative Tracking Accuracy}

To evaluate the closed-loop pointing performance of the YOLO–PID controller, we analyzed the horizontal image-plane error
\[
e_t = x_{\mathrm{bbox,centroid}} - x_{\mathrm{image,center}},
\]
where $x_{\mathrm{bbox,centroid}}$ is the detected object centroid and $x_{\mathrm{image,center}}$ is the image center.  
For interpretability across cameras, this pixel error was converted into an angular deviation
\[
\theta_t[^\circ] = e_t \cdot \mathrm{HFOV}/W,
\]
where $\mathrm{HFOV}$ is the horizontal field of view and $W$ the image width in pixels.  
Frames without a valid detection were marked as \textit{NaN} rather than zero to avoid biasing the statistics.

For each trial, we computed the mean and median absolute error $|e_t|$, the interquartile range, the end-to-end latency from image capture to servo update, and the real-time loop rate (frames per second, FPS) to verify on-board feasibility.

\begin{figure*}[t]
\centering
\begin{subfigure}[b]{0.48\textwidth}
    \includegraphics[width=\linewidth]{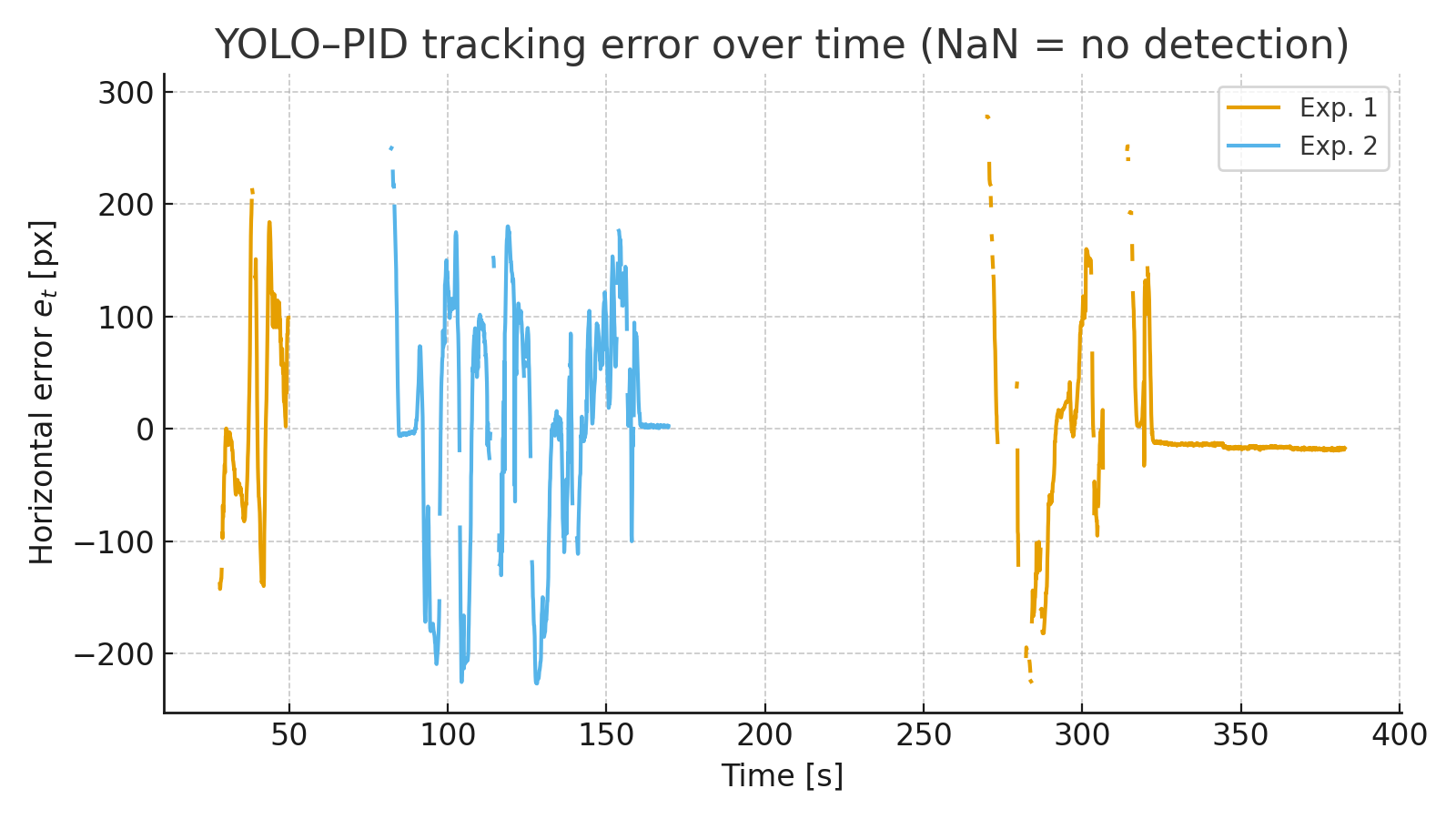}
    \caption*{\footnotesize (a) Horizontal error over time (NaN = no detection).}
\end{subfigure}
\hfill
\begin{subfigure}[b]{0.48\textwidth}
    \includegraphics[width=\linewidth]{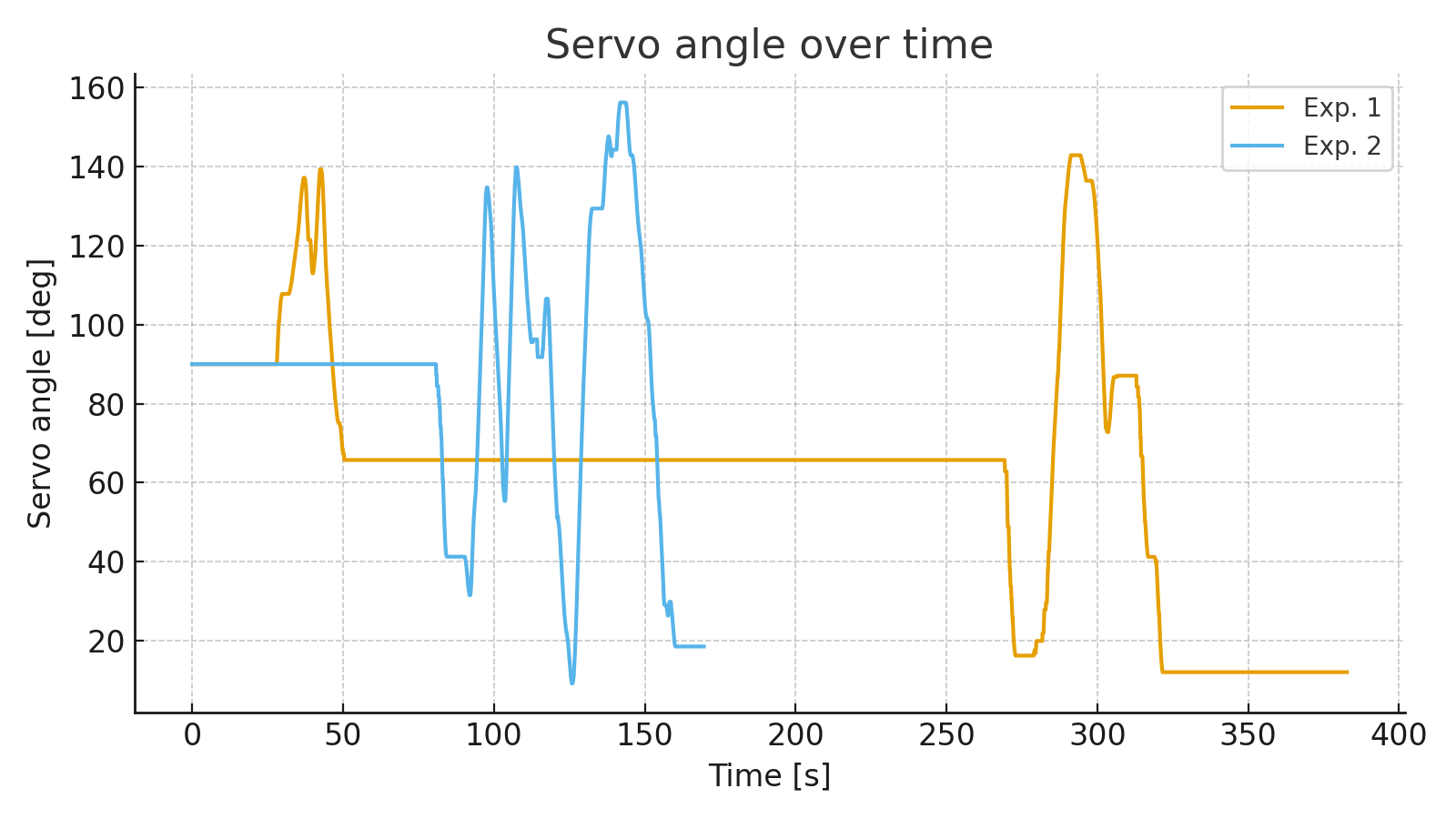}
    \caption*{\footnotesize (b) Servo command evolution.}
\end{subfigure}

\vspace{0.6em}
\begin{subfigure}[b]{0.48\textwidth}
    \includegraphics[width=\linewidth]{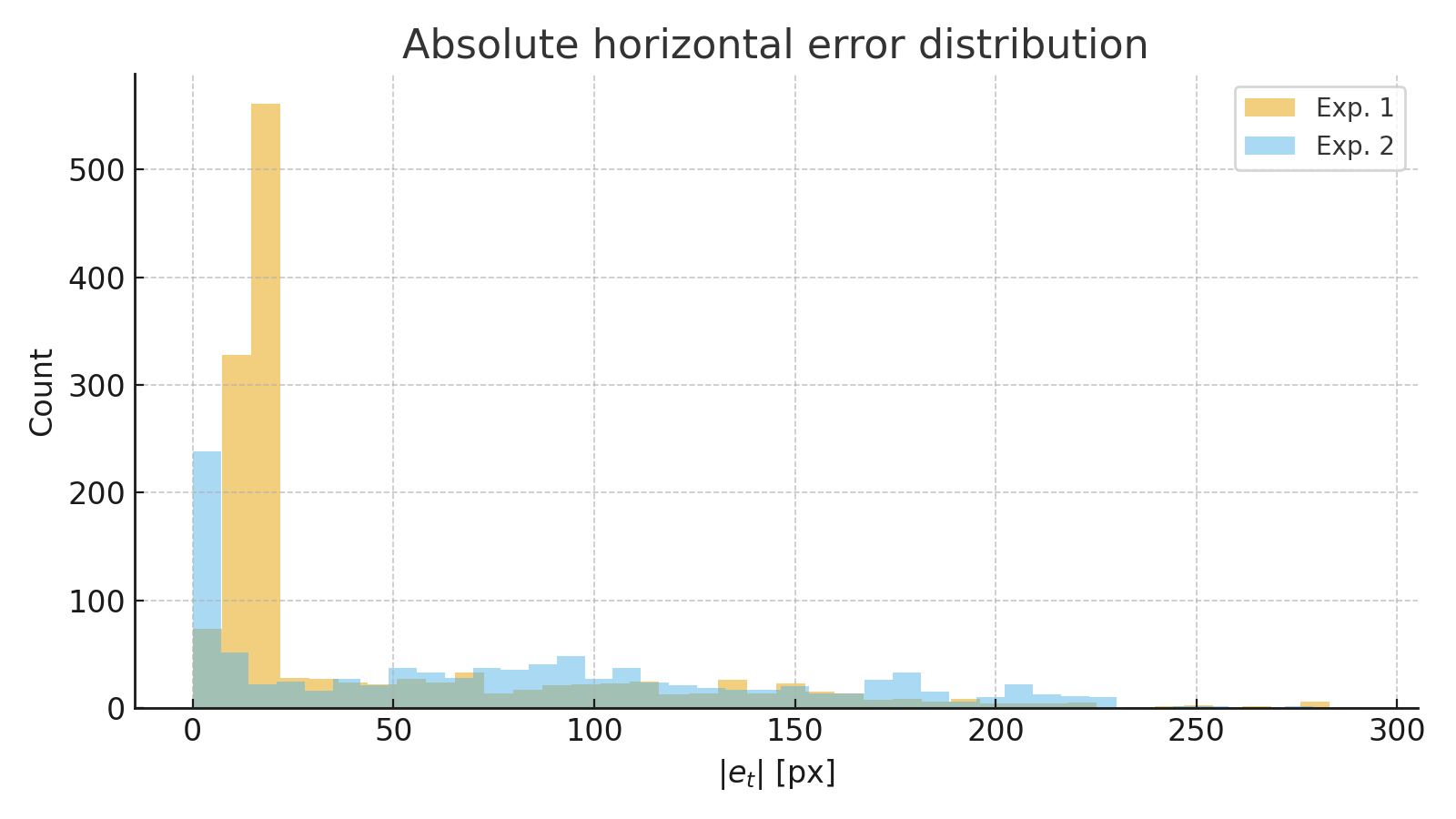}
    \caption*{\footnotesize (c) Distribution of $|e_t|$ across experiments.}
\end{subfigure}
\hfill
\begin{subfigure}[b]{0.48\textwidth}
    \includegraphics[width=\linewidth]{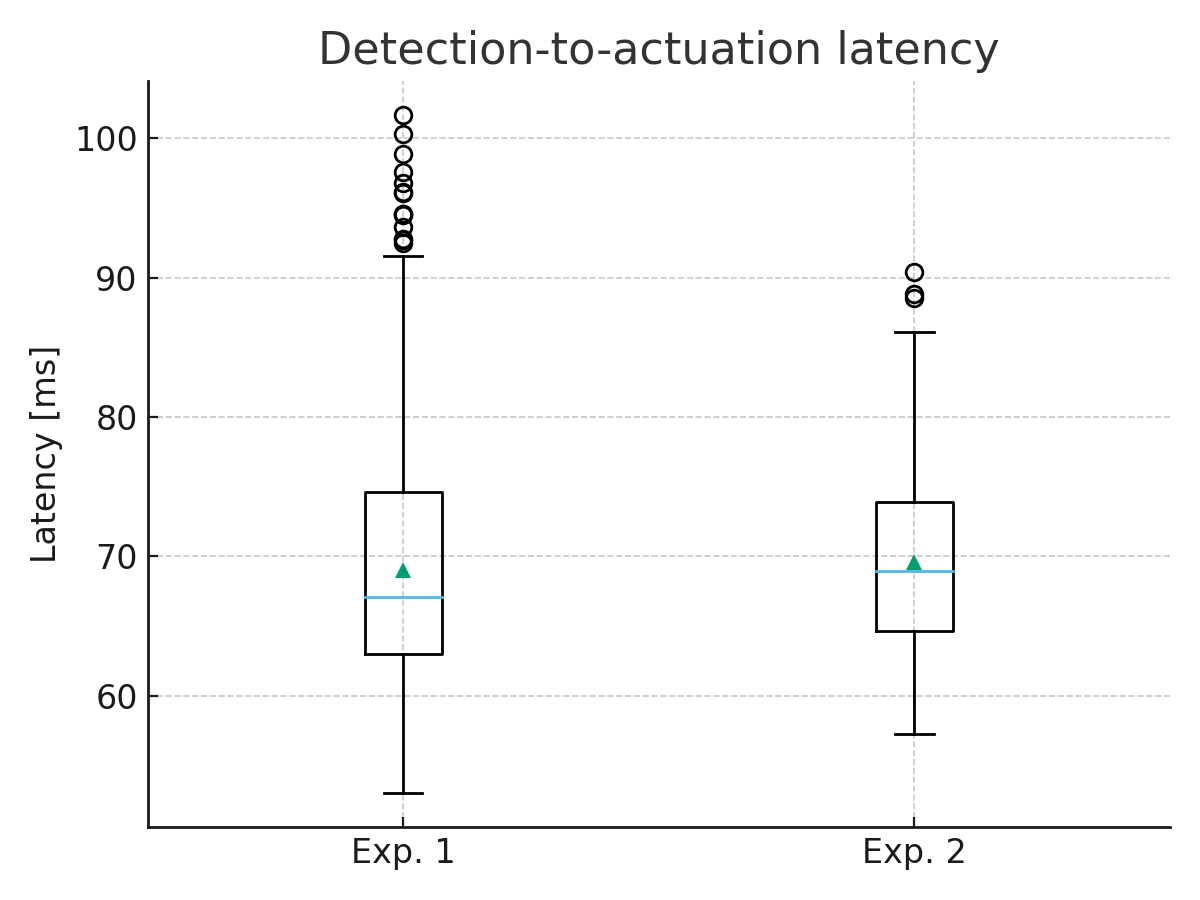}
    \caption*{\footnotesize (d) Detection-to-actuation latency.}
\end{subfigure}
\caption{YOLO–PID tracking metrics from two representative outdoor experiments. Frames without valid detections are excluded (NaN) to prevent artificial zero error values.}
\label{fig:tracking_accuracy}
\end{figure*}

Figure~\ref{fig:tracking_accuracy} summarizes the results. Panel (a) shows the horizontal tracking error over time for two representative outdoor experiments. Gaps indicate frames without a valid detection. Panel (b) presents the servo command evolution, illustrating how the controller responded to changing error dynamics. Panel (c) shows aggregated error distributions, and Panel (d) reports detection-to-actuation latency measured on the embedded platform.

Table~\ref{tab:tracking_metrics} reports the key quantitative metrics. Both experiments achieved real-time operation with mean loop rates above $15$~FPS and median latency near $67$--$69$~ms, sufficient for smooth head movement. Experiment~1, performed under favourable outdoor conditions, achieved a median absolute error of approximately $6$~px (about $0.5^\circ$). Experiment~2, conducted under harsher illumination and more cluttered backgrounds, exhibited lower detection reliability and a slightly higher error spread, with occasional PID overshoot when detections became intermittent. Although detection rates dropped in challenging scenes, the servo loop remained stable and avoided oscillatory failure.

\begin{table}[t]
\centering
\caption{Closed-loop YOLO–PID tracking performance (mean $\pm$ std; median in parentheses).}
\begin{tabular}{lcccc}
\toprule
Trial & FPS & Latency [ms] & Detection rate [\%] & $|e_t|$ [px] \\
\midrule
Exp.~1 & 15.2 $\pm$ 2.2 & 69.0 $\pm$ 8.1 (67.1) & 25.5 & 70.0 $\pm$ 3.0 (15.5) \\
Exp.~2 & 22.0 $\pm$ 7.8 & 69.6 $\pm$ 6.2 (69.0) & 31.7 & 104.0 $\pm$ 7.2 (4.0) \\
\bottomrule
\end{tabular}
\label{tab:tracking_metrics}
\end{table}

These findings confirm that the YOLO–PID loop can operate fully on-board at interactive frame rates with sub-$100$~ms end-to-end latency. Tracking accuracy is primarily constrained by detection confidence under challenging outdoor conditions. Further improvements could be achieved by retraining YOLO on domain-specific Houbara data or incorporating temporal filtering to stabilize detections.

\noindent\textit{Broader applications.}
Once tracking is enabled through the user interface, the perception–action loop runs autonomously and in real time without human intervention. The embedded YOLOv11n detector continuously identifies the target, and the PID controller drives the pan–tilt neck to maintain alignment. All sensing, detection, and actuation are performed locally on the NVIDIA Jetson module at $15$–$22$~FPS with sub-$100$~ms latency. This design constitutes an embodied visual servoing system: perception directly drives physical movement in outdoor environments with natural illumination, sand, and thermal variability, rather than relying on offline or simulation-only control.

\section{Discussion}

This work presents a bio-inspired robotic platform that combines anatomical fidelity, modular digital fabrication, and real-time perception to support both scientific inquiry and public engagement. The female Houbara bustard surrogate shows how engineering precision and biomimetic design can produce a field-capable system that elicits ecologically meaningful behaviors from live animals while remaining suitable for controlled experimentation and outreach. Its lifelike morphology enables natural recognition by conspecifics, and the embedded perception–action loop supports repeatable behavioral trials under challenging desert conditions.

Positioned within the emerging field of expressive and ecologically integrated robots, HuBot demonstrates how biomimetic intelligence can extend beyond controlled laboratory settings toward robust, deployable systems. By uniting high-fidelity morphology with programmable autonomy, it advances animal–robot interaction research, contributes to conservation technologies, and illustrates how bio-inspired agents can serve simultaneously as scientific instruments and platforms for public engagement. Such systems bridge the rigour of ecological research with the narrative potential of kinetic art and environmental storytelling.

To contextualise HuBot within current bio-inspired field robotics, Table~\ref{tab:bio_robots_comparison} compares representative platforms reported between 2020 and 2025. Most existing systems emphasise mobility or basic sensing but lack ecological validation and rarely integrate thermal or night-vision perception. For example, Robo Car~\cite{savale2024ecological} applies CNN-based object detection for terrestrial monitoring but has only been evaluated in forested areas; Robotic Fish~\cite{manduca2023maneuverability} and spider-inspired systems~\cite{singh2025bioinspired} explore adaptive control but remain largely laboratory-focused; and quadruped or climbing robots~\cite{suarez2019quadruped,chattopadhyay2018adhesion} are primarily tested indoors. 

HuBot differs by combining \emph{photorealistic avian morphology}, a \emph{rocker–bogie chassis} tailored for desert terrain, and a \emph{multi-modal RGB–thermal sensing suite} with real-time YOLO-based detection and embodied PID visual servoing. Importantly, it has been validated with live Houbara bustards in natural desert conditions, demonstrating both ecological acceptance and technical robustness. The inclusion of thermal–visible fusion through the NightFusion module aligns with the broader trend toward lightweight, interpretable, and field-ready AI-driven perception highlighted in recent advances in biomimetic bird robots and field-deployable artificial intelligence.

\begin{table*}[t]
\centering
\caption{Comparison of representative bio-inspired field robots (2020--2025) with the proposed HuBot platform.}
\label{tab:bio_robots_comparison}
\begin{tabular}{p{2.5cm} p{2.1cm} p{1.8cm} p{2.6cm} p{1.8cm} p{1.3cm} p{1.3cm}}
\toprule
\textbf{Robot} & \textbf{Species} & \textbf{Mobility} & \textbf{Sensing} & \textbf{Control} & \textbf{Field} & \textbf{Thermal} \\
\midrule
Robo Car~\cite{savale2024ecological} & N/A & Wheeled & RGB + CNN & ConvNeXt & \cmark & \xmark \\
Robotic Fish~\cite{manduca2023maneuverability} & Fish & Swimming & Proprioceptive & CPGs & \xmark & \xmark \\
Spider robot~\cite{singh2025bioinspired} & Spider & Quad/Hexa & N/A & PD/PID & \xmark & \xmark \\
Quadruped~\cite{suarez2019quadruped} & Quadruped & Walking & CO$_2$ sensors & N/A & \xmark & \xmark \\
Climbing robots~\cite{chattopadhyay2018adhesion} & Various & Climbing & Intelligent sensors & N/A & \xmark & \xmark \\
\textbf{HuBot (ours)} & \textbf{Houbara bustard} & \textbf{6-wheel rocker--bogie} & \textbf{RGB + Thermal} & \textbf{YOLOv11 + PID} & \textbf{\cmark} & \textbf{\cmark} \\
\bottomrule
\end{tabular}
\end{table*}

This comparison highlights HuBot’s contribution to biomimetic field robotics. It bridges the gap between high-fidelity animal morphology and durable autonomous operation in harsh outdoor environments. Whereas many prior systems remain proof-of-concept or simulation-bound, HuBot demonstrates ecological validity, modular construction suitable for long-term deployment, and AI-driven perception that enables real-time, low-latency interaction with wildlife.

\section{Limitations}

Although the platform proved mechanically and perceptually robust in desert field trials, several constraints remain. Locomotion is currently wheel based, which limits the replication of natural avian gait and postural dynamics that influence complex courtship and territorial displays. The absence of articulated lower limbs also reduces manoeuvrability on irregular terrain compared with legged or hybrid chassis.

Perception is restricted to RGB and thermal imaging, with the thermal module validated only against surrogate warm-bodied subjects. The system cannot yet interpret subtle thermal signatures, fine-scale proximity cues, or social signals from conspecifics and human observers. Autonomy remains modest: interaction is largely operator-triggered, and adaptive behaviors are limited compared with advanced laboratory bio-robotic systems. Finally, endurance is constrained by bulky battery packs, which restrict long-duration untethered operation in remote habitats.

\section{Future Work}

Future developments will focus on increasing ecological realism, autonomy, and operational endurance. A legged chassis with articulated joints is being designed to support biologically plausible locomotion and interactive behaviors such as approach, retreat, and lateral display. Perception will be enhanced with ambient light and proximity sensing, together with an upgraded thermal module capable of accurately mapping body temperature in live birds. These improvements aim to enable context-aware responses to environmental stimuli and conspecific presence.

Control strategies will evolve beyond the current fixed-gain visual servoing toward adaptive and learning-based approaches. Achieving fully adaptive visual servoing in uncontrolled outdoor environments remains challenging because of real-time latency constraints and noisy detections. Although the present system demonstrates embodied, closed-loop visual tracking, future work will explore adaptive gain tuning and reinforcement learning to improve robustness under clutter, occlusion, and variable lighting. Simulation-to-reality transfer will be employed to pre-train control policies and increase stability before field deployment.

Energy management will also be addressed through improved battery technology and the exploration of lightweight solar-assisted charging to extend untethered operation in remote habitats. Integration of GPS, inertial measurement units, and long-range wireless communication will support semi-autonomous navigation and coordinated multi-agent studies across wide territories. Acoustic playback and context-aware vocalisation will be investigated to enhance behavioral salience during courtship and territorial interactions. In parallel, user-friendly interfaces will be developed to support ecological research, conservation training, and educational outreach. All upgrades will undergo structured validation with live Houbara bustards to ensure ecological acceptance and reliable operation under natural field conditions.

\section{Conclusion}

This work presented a modular framework for creating lifelike, terrain-capable robotic surrogates, demonstrated through the development of a bio-inspired female Houbara bustard. The workflow integrates high-resolution 3D scanning, parametric CAD modeling, articulated 3D printing, and UV-textured vinyl wrapping to achieve structural accuracy and visual fidelity suitable for both ecological research and public engagement.

The resulting platform combines an NVIDIA Jetson-based onboard system with real-time perception, RGB and thermal sensing for day and night operation, and embodied PID visual servoing for interactive engagement. Field trials confirmed its ability to elicit natural behavioral responses from live Houbara while maintaining mechanical robustness and sensory reliability under harsh desert conditions.

Although the current system is limited by wheeled locomotion, fixed-gain visual servoing, and moderate energy autonomy, it provides a reproducible and extensible foundation for future bio-inspired field robots. Planned enhancements include legged mobility, adaptive learning-based control, extended endurance through improved power management, and richer multimodal sensing for context-aware interaction.

Beyond ecological and behavioral studies, the proposed approach offers a transferable blueprint for designing expressive, field-ready bio-inspired robots that unite scientific functionality with public-facing impact. This cross-disciplinary integration of robotics, ecology, and biomimetic intelligence advances both rigorous field experimentation and accessible science communication, contributing to the emerging class of intelligent, ecologically integrated robotic systems.

\section*{Ethics statement}
All procedures involving live Houbara bustards were conducted under the approval of the National Avian Research Center and complied with local wildlife protection regulations. Experimental protocols followed the ethical standards of animal research and were designed to minimize disturbance and stress.

\section*{Project Resources}

To facilitate reproducibility and further development, we provide open access to the fabrication pipeline, control software, and demonstration videos of the proposed robotic platform. All materials, including code, 3D models, and supplementary field recordings, are available at:
\href{https://lyessaadsaoud.github.io/KineBot/}{\texttt{https://lyessaadsaoud.github.io/KineBot/}}

\section*{Author Declaration on the Use of AI}

Language editing and text refinement were partially assisted by AI tools. All content was reviewed and approved by the authors, who take full responsibility for the final manuscript.

\section*{Acknowledgment}
This work is supported by Khalifa University, United Arab Emirates under award Hubot: Houbara robot for behavioral studies in the field and sampling 8434000544.

\bibliographystyle{unsrt}
\bibliography{reference}

\end{document}